\documentclass[sigconf, nonacm]{acmart}
\usepackage{hyperref}
\usepackage{url}
\newcommand{\proposed}{\textsf{TGLRN}}
\usepackage{tablefootnote}
\usepackage{multirow}
\newtheorem{definition}{Definition}
\usepackage[normalem]{ulem}
\useunder{\uline}{\ul}{}
\usepackage{enumitem}
\usepackage{array}
\usepackage{caption}
\usepackage{multirow}
\usepackage{graphicx}
\usepackage{wrapfig}
\usepackage{booktabs}
\usepackage{adjustbox}
\usepackage{lineno}
\usepackage{algorithm} 
\usepackage{algpseudocode}
\usepackage{placeins}

\usepackage{pifont}
\newcommand{\cmark}{\ding{51}}%
\newcommand{\xmark}{\ding{55}}%

\usepackage{wrapfig}
\usepackage{xcolor}

\setcopyright{none}
\renewcommand\footnotetextcopyrightpermission[1]{} 

\AtBeginDocument{%
  \providecommand\BibTeX{{%
    \normalfont B\kern-0.5em{\scshape i\kern-0.25em b}\kern-0.8em\TeX}}}

\begin{document}

\title[Temporal Graph Learning Recurrent Neural Network for Traffic Forecasting]{Temporal Graph Learning Recurrent Neural Network\\for Traffic Forecasting}

\author{Sanghyun Lee}
\affiliation{%
  \institution{KAIST AI}
  \country{}
  }
\email{ragdoll1762@kaist.ac.kr}

\author{Chanyoung Park}
\affiliation{%
  \institution{KAIST ISysE \& AI}
  \country{}
}
\email{cy.park@kaist.ac.kr}

\begin{abstract}
Accurate traffic flow forecasting is a crucial research topic in transportation management. However, it is a challenging problem due to rapidly changing traffic conditions, high nonlinearity of traffic flow, and complex spatial and temporal correlations of road networks. 
Most existing studies either try to capture the spatial dependencies between roads using the same semantic graph over different time steps, or assume all sensors on the roads are equally likely to be connected regardless of the distance between them. However, we observe that the spatial dependencies between roads indeed change over time, and two distant roads are not likely to be helpful to each other when predicting the traffic flow, both of which limit the performance of existing studies.
In this paper, we propose Temporal Graph Learning Recurrent Neural Network (\proposed) to address these problems. 
More precisely, to effectively model the nature of time series, we leverage Recurrent Neural Networks (RNNs) to dynamically construct a graph at each time step, thereby capturing the time-evolving spatial dependencies between roads (i.e., microscopic view). Simultaneously, we provide the Adaptive Structure Information to the model, ensuring that close and consecutive sensors are considered to be more important for predicting the traffic flow (i.e., macroscopic view).
Furthermore, to endow~\proposed~with robustness, we introduce an edge sampling strategy when constructing the graph at each time step, which eventually leads to further improvements on the model performance.
Experimental results on four commonly used real-world benchmark datasets show the effectiveness of~\proposed.
\end{abstract}

\settopmatter{printfolios=true}
\maketitle

\section{Introduction}
\label{sec:intorduction}
Research on intelligent transport systems has been actively conducted recently as 
traffic information becomes easily accessible thanks to real-time collection and management technology.
Accurate traffic flow forecasting based on historically observed data is a crucial part in intelligent transport systems, which makes it possible to maximize the use of transportation facilities, increase transportation efficiency, and promote the convenience and safety of people. However,  accurate forecasting of future traffic flows based on historical data is challenging because it is necessary to simultaneously capture the complex spatial dependencies and temporal dynamics of road networks.

\begin{figure}[htbp]
\centerline{\includegraphics[width=1\linewidth]{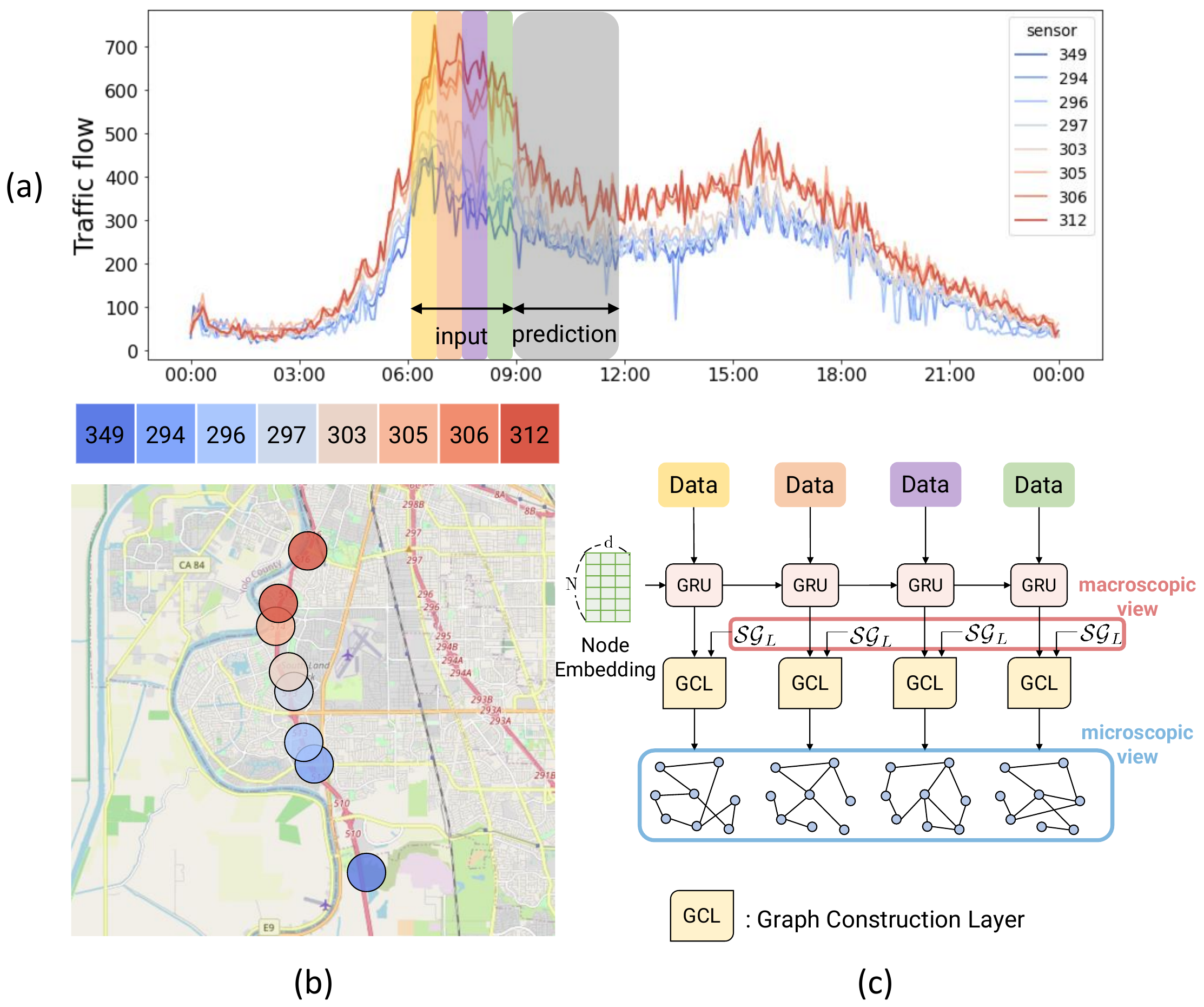}}
\caption{(a) Traffic flow, and (b) sensor locations that exist consecutively. From a macroscopic view, consecutive sensors exhibit similar trends (upward or downward) due to spatial proximity. (c) At the microscopic level, we employ RNNs to model the dynamic spatial dependencies between sensor pairs, constructing different weighted graphs for each time step. 
}
\label{fig1}
\end{figure}

For traffic forecasting, traditional statistical machine learning models such as historical average (HA), autoregressive integrated moving average (ARIMA)~\cite{arima}, and support vector regression (SVR)~\cite{svr} have been widely used. However, as they fall short of capturing complex spatio-temporal relationships between roads, researchers have been recently adopting deep learning techniques~\cite{tedjopurnomo2020survey}.
Specifically, temporal dependencies have been modeled using recurrent neural networks (RNN)~\cite{dcrnn,digc-net,mra-bgcn,m-rgnn,zgcnnet,agcrn,gaan}. Additionally, some studies have explored the use of convolutional neural networks (CNN)~\cite{stgcn,astgcn,wavenet,stag-gcn,hgcn,lsgcn,stfgnn,stgode} and the attention mechanism~\cite{st-gart,gman,dstagnn,astgcn} for modeling temporal dependency.
Besides, as roads that are geographically adjacent to each other
have high spatial dependencies, a road network can be expressed based on a graph whose nodes denote sensors on the roads and edges denote the connections between the sensors where the edge weights are determined by the distance between the sensors.
With a graph constructed in this way, spatial dependencies can be modeled using graph convolution networks (GCN)~\cite{gcn,gcn2,gcn3,gat}. However, since they rely on a graph constructed with adjacent roads, it is impossible to learn spatial dependencies between geographically distant nodes. To address this problem and discover the hidden spatial dependencies between roads, there are studies that use learned graph structures~\cite{wavenet,gts,stemgnn}, graph Laplacian~\cite{agcrn,zgcnnet}, or dynamic time warping (DTW)~\cite{stfgnn,stgode,stag-gcn}.

Although current deep learning-based methods have shown effectiveness in forecasting traffic flow by capturing both spatial and temporal dependencies, they are constrained in two significant aspects: First, existing methods often overlook the \textit{dynamic nature of spatial relationships among roads, which change in response to real-time conditions from a microscopic perspective.} As depicted in Figure \ref{fig1}, to accurately reflect the evolving spatial dependencies of traffic flow from a microscopic viewpoint, we could consider modeling that constructs more refined graphs at each time step.  Typically, existing approaches utilize the same semantic information graph
at every time step, lacking the capability to adjust the structures of spatial dependencies dynamically at each time step. 
While some methods attempt to capture these Dynamic Time Graphs \cite{dstagnn,dynamic_time_1,dynamic_time_2,dynamic_time_3,dynamic_time_4}, they differ significantly from our model for the following reasons.

\begin{figure}[t]
\centerline{\includegraphics[width=0.9\linewidth]{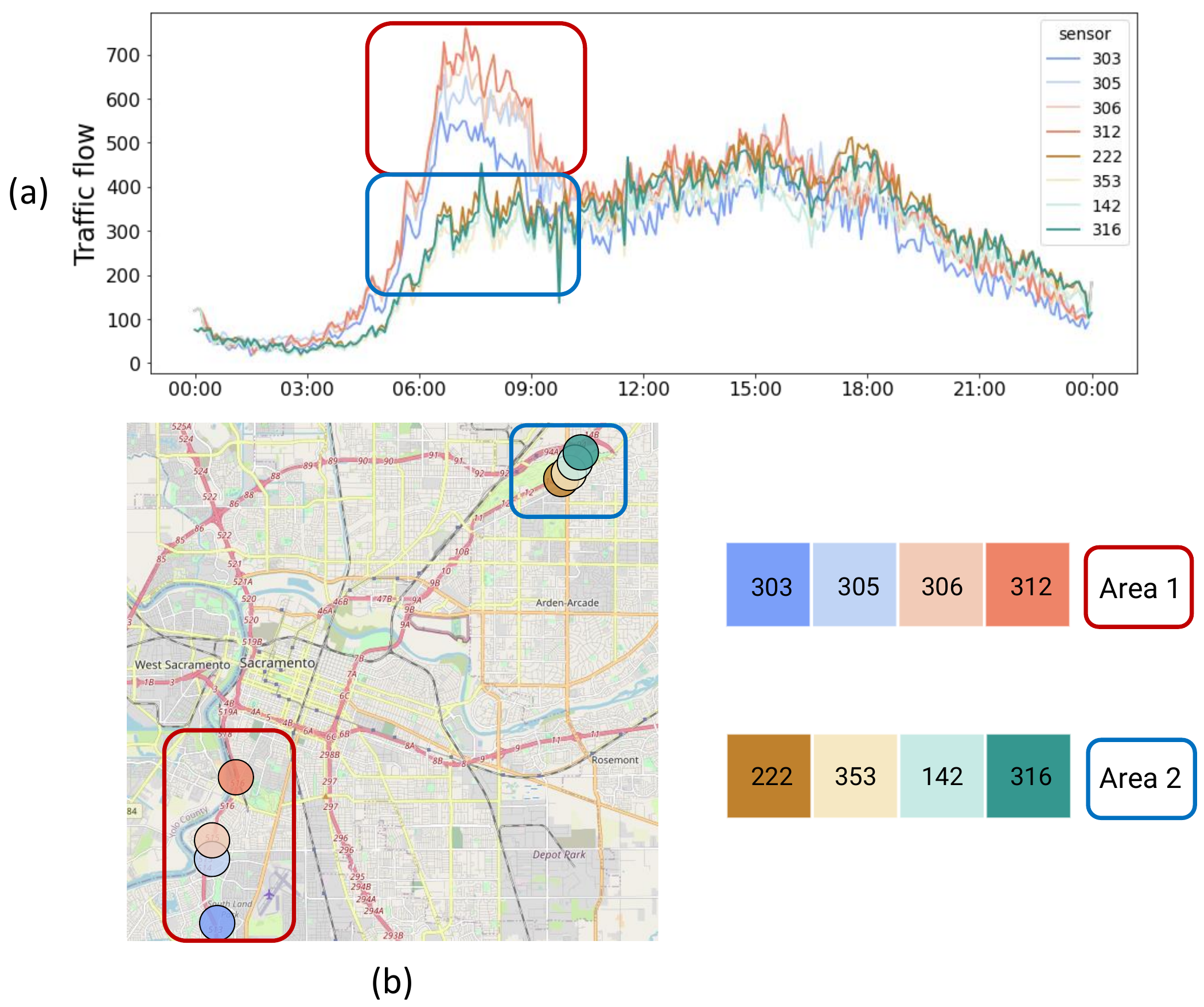}}
\caption{(a) Traffic flow, and (b) location of consecutive sensors in two geographically distant areas. The two geographically distant areas exhibit distinct trends. Hence, we construct a graph where road pairs are likely to be connected if the roads are within a certain distance.
}
\label{fig2}
\vspace{-1ex}
\end{figure}

Second, existing methods 
assume that \textit{all sensors on the roads are equally likely to be connected by edges (even though the weights may differ) regardless of the distance between them.}
More precisely, methods that rely on the learned graph structures, and graph Laplacian construct graphs in which every pair of nodes is connected (i.e., complete graph). 
Moreover, methods that rely on DTW constructs graphs in which two nodes are connected solely based on the similarity between their traffic flow values, regardless of the distance between them.
Existing methods that adopt Dynamic Time Graphs \cite{dstagnn,dynamic_time_1,dynamic_time_2,dynamic_time_3,dynamic_time_4} also do not consider the relative distance of the sensors. 
However, two distant roads are not likely to be helpful to each other when predicting the traffic flow of each road, because, in general, roads in geographically distant areas have different trends, as shown in Figure ~\ref{fig2}.
Even if two roads in geographically distant areas have shown similar patterns in the past, when an event (e.g.,  traffic jam or local event) occurs on one road, this road would exhibit a completely different pattern from the other road as these two roads are distant from each other.
Hence, we argue that the strategy for learning graph structures should involve a targeted approach, focusing on geographically proximate sensors positioned consecutively (i.e., macroscopic view), instead of incorporating every sensor available.

Furthermore, in dynamic environments like traffic networks, \textit{robustness in prediction is paramount}. Consider the myriad of unpredictable events that can influence traffic, such as sudden changes in weather, unplanned roadworks, or unforeseen public events. Such unpredictable factors can drastically change the traffic patterns on roads. Even subtle changes in external factors can ripple through the traffic network, causing significant and sometimes surprising changes in flow and congestion. These myriad variations highlight the need for models that are not only accurate but also robust. A model that is too rigid or overly tailored to past patterns may fail to adapt to new and unexpected scenarios, leading to suboptimal predictions. The inherent unpredictability and complexity of traffic patterns further underscore the necessity for robust prediction mechanisms in traffic forecasting.

\begin{figure*}[t]
\begin{center}
\includegraphics[width=0.85\linewidth]{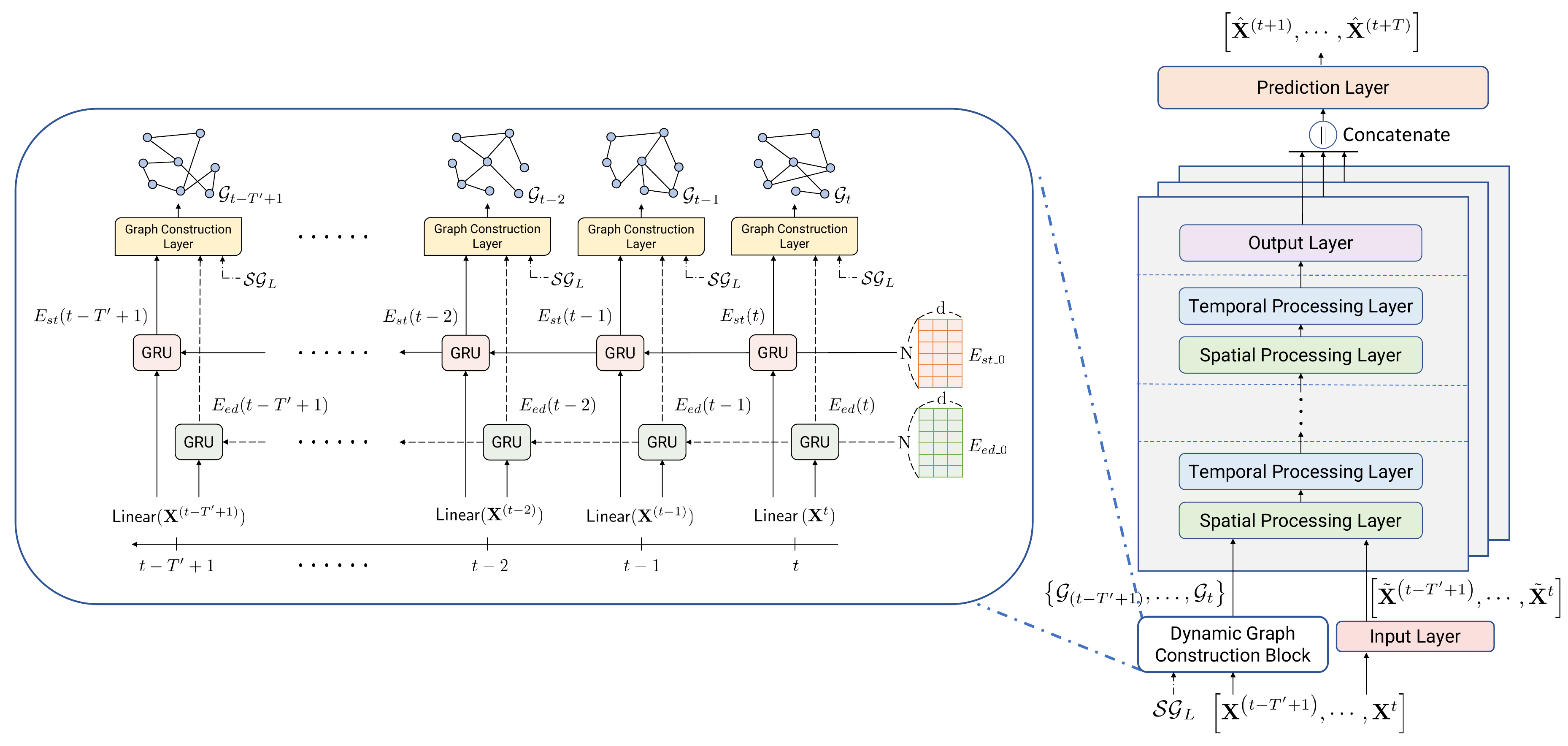}
\end{center}
\vspace{-1ex}
\caption{Overall architecture of~\proposed.}
\label{fig3}
\end{figure*}

In the light of the aforementioned limitations, we propose Temporal Graph Learning Recurrent Neural Network (\proposed) for traffic flow forecasting. Unlike existing models,~\proposed~dynamically constructs an input graph at each time step by introducing node embeddings that change over time (i.e., \textit{microscopic view}).
More precisely, to effectively model the nature of time series, we dynamically compute the hidden state of each node using the RNN structure at each time step, and construct dynamic graphs using the time-varying node embeddings (Refer to Figure~\ref{fig1}(c)).
Moreover, we introduce the Adaptive Structure Information to the model in which graphs are constructed by considering the relative geographical distance between the nodes (i.e., macroscopic view). 
Specifically, we provide the model another graph in which consecutive nodes that are located within a certain geographical distance from one another exhibit high spatial dependencies (Refer to Figure~\ref{fig1} and Figure ~\ref{fig2}).

{Furthermore, to endow~\proposed~with robustness, we introduce an edge sampling strategy when constructing the graph at each time step, which eventually leads to further improvements on the model performance.} Through extensive experiments on four commonly used real-world benchmark datasets, we demonstrate that~\proposed~shows superior performance compared with state-of-the-art models.

\smallskip
\section{Related Work}
\smallskip
\subsection{Traffic Flow Forecasting}
\noindent{\textbf{Classical and Deep learning-based approaches. }}
Classical statistical machine learning models used for traffic forecasting include HA, ARIMA\cite{arima}, KNN, and SVR\cite{svr}. Although these methods achieved promising results, these models make predictions solely based on temporal data from previous time steps, ignoring spatial dependencies between roads. Recent deep learning-based approaches capture complex spatio-temporal dependencies and show more accurate prediction performance. More precisely, \cite{rnncnn,rnncnn2,rnncnn3} learn temporal features using RNN and spatial features using CNN both of which facilitate simultaneous modeling of complex spatio-temporal dependencies. 
However, since the above models are primarily designed for grid-structured data, adapting them to road networks is not straightforward.

\smallskip
\noindent{\textbf{Graph-based approaches. }}
In order to address the above challenge, recent studies adopt GCN~\cite{gcn} to perform spatial-temporal forecasting. More precisely, DCRNN~\cite{dcrnn} models the spatial dependencies between roads by applying a diffusion process to the graph structure, and the temporal dependencies using the sequence to sequence architecture based on RNN. STGCN \cite{stgcn} proposes learning spatio-temporal dependencies using GCN and temporal convolution. ASTGCN~\cite{astgcn} captures dynamic spatial and temporal correlations using spatial-temporal attention in graph convolution and temporal convolution.
However, as these models rely on the geographic distance between roads to construct graphs, they overlook the spatial dependencies between distant roads despite possible dependencies therein.
Hence, in order to capture spatial dependencies that cannot be captured using geographical distance, Graph Wavenet~\cite{wavenet} learns the adjacency matrix through learnable node embeddings, and AGCRN~\cite{agcrn} learns graph Laplacian through learnable node embeddings. Furthermore, to model Dynamic Time Graphs that change over time, methodologies such as \cite{dstagnn,dynamic_time_1,dynamic_time_2,dynamic_time_3,dynamic_time_4} exist.

\begin{figure*}[t]
\begin{center}
\includegraphics[width=0.8\linewidth]{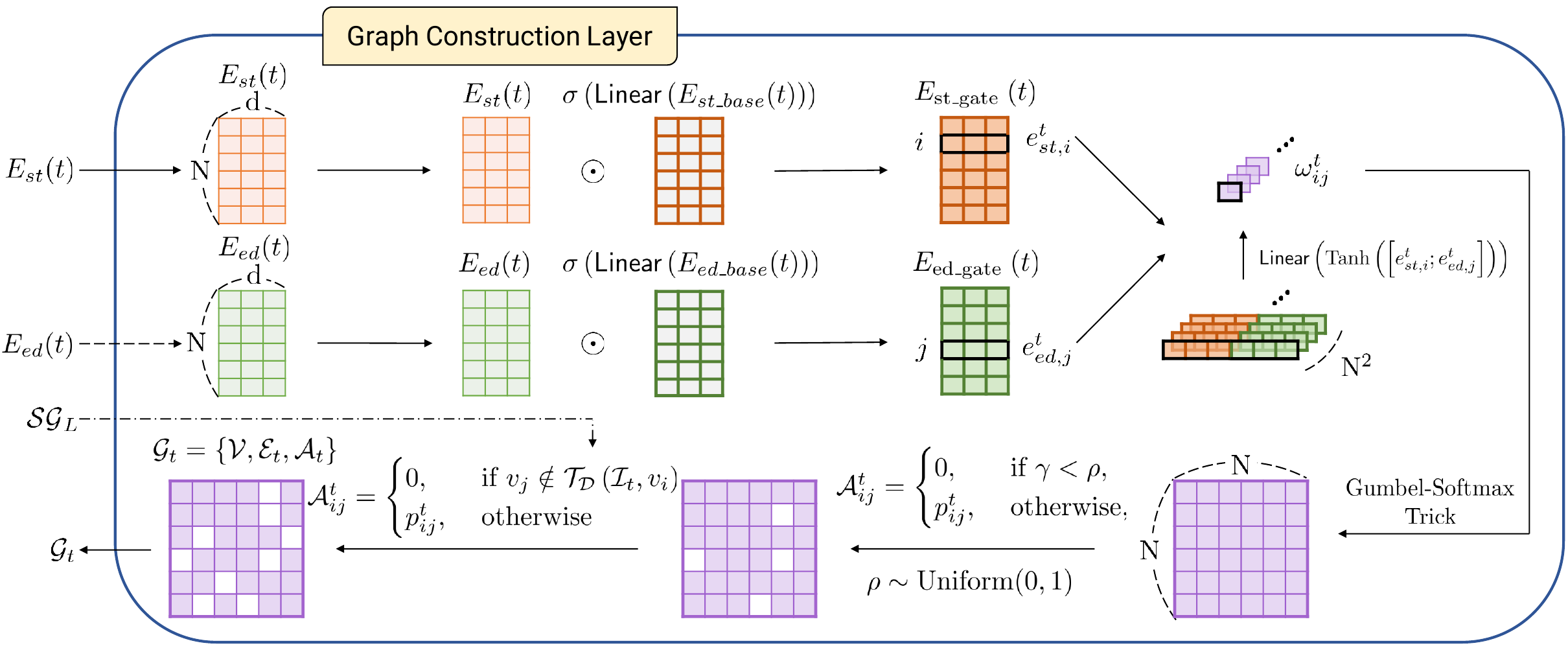}
\end{center}
\caption{Overall architecture of Graph Construction Layer (GCL).}
\label{fig:GSL}
\end{figure*}

\smallskip
\section{Preliminaries}
\smallskip
\subsection{Problem Definition}
\label{sec:PD}
We define a road network as a directed graph $\mathcal{G}=\{\mathcal{V}, \mathcal{E},\mathcal{A}\}$, where  $\mathcal{V}$ is the set of nodes with $N=\left| \mathcal{V} \right| $ traffic sensors, $\mathcal{E} \subseteq \mathcal{V} \times \mathcal{V}$ is the set of edges, and $\mathcal{A}\in \mathbb{R}^{N \times N}$ is an adjacency matrix of $\mathcal{G}$. In general, $\mathcal{G}$ is constructed based on the relationship between spatial neighbors, and it can be denoted by $\mathcal{G}^{s p}=\{\mathcal{V}, \mathcal{E}^{s p},\mathcal{A}^{s p}\}$, where $\mathcal{A}^{s p}\in \mathbb{R}^{N \times N}$ 
connects consecutively existing sensors of the actual road network:

\begin{equation}
\mathcal{A}_{i j}^{s p}= \begin{cases}1, &\text {if sensors } i \text { and } j \text { exist consecutively,} \\ 0, & \text {otherwise. }\end{cases}
\label{eq1}
\end{equation}

In this research, we define `consecutively existing' sensors using two distinct and separate configurations across datasets. The primary configuration aligns with the sequence of travel on a roadway: a sensor is considered `consecutive' to another if it is the immediate next sensor encountered along the travel path. Specifically, sensor $j$ is `consecutive' to sensor $i$ if a vehicle passes by sensor $i$ first and then sensor $j$. The secondary configuration is predicated on spatial proximity, where `consecutively existing' refers to a small, predefined number of sensors that are geographically nearest to a reference sensor.

The traffic flow data observed on node $i$ at time $t$ is denoted by $\mathbf{x}_{t}^{i} \in \mathbb{R}^{F}$, where $F$ is the number of features. $\mathbf{X}^{(t)}= \left(\mathbf{x}_{t}^{1}, \mathbf{x}_{t}^{2}, \ldots, \mathbf{x}_{t}^{N}\right) \in \mathbb{R}^{N \times F}$ denotes the traffic flow matrix of the entire road network at time step $t$. The problem of traffic forecasting is defined as learning a function $\mathcal{F}$ to forecast the next $T$ time steps based on data collected for previous $T^{\prime}$ time steps, and the road network $\mathcal{G}$:

\begin{equation}
\left[\mathbf{X}^{\left(t-T^{\prime}+1\right): t}, \mathcal{G}\right] \stackrel{\mathcal{F}}{\rightarrow}\left[\mathbf{X}^{(t+1):(t+T)}\right].
\label{eq:prob1}
\end{equation}
\smallskip
\subsection{Dynamically Changing Graph Structures}
We are interested in graph structures that dynamically change over time. Let $\mathcal{G}_{t}=\{\mathcal{V}, \mathcal{E}_{t},\mathcal{A}_{t}\}$ represent the directed road network at time step $t$. The sequence of dynamically changing graph structures over time is denoted by $\left\{\mathcal{G}_{\left(t-T^{\prime}+1\right)}, \ldots, \mathcal{G}_{t}\right\}$. 
Then, the problem of traffic forecasting defined in Equation~\ref{eq:prob1} is reformulated as follows:
\begin{equation}
\left[\mathbf{X}^{\left(t-T^{\prime}+1\right): t}, \mathcal{G}_{\left(t-T^{\prime}+1\right): t}\right] \stackrel{\mathcal{F}}{\rightarrow}\left[\mathbf{X}^{(t+1):(t+T)}\right].
\label{eq:prob2}
\end{equation}

\smallskip
\section{Proposed Method:~\proposed}
\smallskip
\subsection{Model Overview}
In this section, we begin by explaining the input layer (\textbf{Section~\ref{sec:IL}}) followed by the two main components, i.e., Dynamic Graph Construction Block \textbf{(Section~\ref{sec:DGCB})} and Spatio-Temporal Block \textbf{(Section~\ref{sec:STB})}. Lastly, we introduce the prediction layer (\textbf{Section~\ref{sec:PL}}) and the final loss function (\textbf{Section~\ref{sec:LF}}). Figure~\ref{fig3} shows the overall architecture of the Temporal Graph Learning Recurrent Neural Network (\proposed).

\smallskip
\subsection{Input Layer}
\label{sec:IL}
The input layer consists of a single linear layer without non-linearity. This layer is responsible for transforming the feature dimension of the input data into a higher dimensional space. 
That is, the input data $\mathbf{X}^{(t)}\in\mathbb{R}^{N\times F}$ is transformed into $\tilde{\mathbf{X}}^{(t)}\in\mathbb{R}^{N\times D}$,
where $F<D$.

\smallskip
\subsection{Dynamic Graph Construction Block}
\label{sec:DGCB}
\smallskip
\subsubsection{Dynamic Node Embedding}
\label{sec:DGCB_DNE}
As described in Section \ref{sec:intorduction}, our objective is to model the evolving spatial dependencies among nodes within a road network accurately. To capture such spatial dependencies, it is necessary to construct different graphs over time, and our goal is to learn the graph structure at each time step. As shown in Figure~\ref{fig3}, we first define two initial node embeddings $E_{s t\_0} \in \mathbb{R}^{N \times d}$ and $E_{e d\_0}\in \mathbb{R}^{N \times d}$ for constructing a graph, where $N$ and $d$ denote the number of nodes and the dimension size of node embeddings, respectively. By using the RNN structure~\cite{lstm,gru}, we obtain intermediate dynamics of hidden states $E_{s t}(t) \in \mathbb{R}^{N \times d}$ and $E_{e d}(t) \in \mathbb{R}^{N \times d}$ (i.e., node embedding) of nodes as follows:
\begin{equation}
\begin{aligned}
&z=\sigma\left(f_z\left(\left[E_{*}(t) ; \textsf { Linear}\left(\mathbf{X}^{(t-1)}\right)\right]\right)\right), \\
&r=\sigma\left(f_r\left(\left[E_{*}(t); \textsf { Linear}\left(\mathbf{X}^{(t-1)}\right)\right]\right)\right), \\
&h^{\prime}=g\left(\left[r \odot E_{*}(t) ; \textsf { Linear}\left(\mathbf{X}^{(t-1)}\right)\right]\right), \\
&E_{*}(t-1)=(1-z) \odot h^{\prime}+z \odot E_{*}(t),
\end{aligned}
\end{equation}
where $f_z$ and $f_r$ are neural networks, $\odot$ denotes the Hadamard product, $\sigma$ denotes sigmoid activation function, $*$ can be either $st$ or $ed$. Note that given the current time step $t$, we run through the past. i.e., the next time step is $t-1$.
We can simplify the above equations as:
\begin{equation}
E_{*}(t-1) = \textsf {GRU-Cell}({E}_{*}(t), \mathbf{X}^{(t-1)}).
\label{eq:grucell}
\end{equation}

In the formulation of our Dynamic Node Embedding model, we establish two separate embeddings, $E_{st}(t)$ and $E_{ed}(t)$, to distinctly address the directional aspects of traffic flows within the network.  This separation is critical for the implementation of the Diffusion Convolution Operation~\cite{dcrnn}—detailed in Section~\ref{sec:STB_SPL}—which relies on a directional graph to effectively model traffic flows. The necessity to differentiate traffic direction - distinguishing upstream from downstream patterns - is crucial for this operation and motivates our use of $E_{st}(t)$ and $E_{ed}(t)$ to construct a graph that inherently reflects directional influences, thereby enhancing the accuracy of the temporal-spatial dependency representation in dynamic road networks.

\smallskip
\subsubsection{Graph Construction Layer}
\label{sec:gcl}
In this section, we aim to construct graph $\mathcal{G}_{t}=\{\mathcal{V}, \mathcal{E}_{t},\mathcal{A}_{t}\}$ for each time step $t$. As shown in Figure~\ref{fig:GSL}, at each time step $t$, the Graph Construction Layer has learnable node embeddings $E_{st\_base}(t)$ and $E_{ed\_base}(t)$ with the same dimensions as $E_{st}(t)$ and $E_{ed}(t)$, and $E_{st\_base}(t)$ and $E_{ed\_base}(t)$ are used for gating. Node embeddings after gating are computed as follows:
\begin{equation}
E_{*\_gate}(t)=E_*(t) \odot\sigma\left(\textsf { Linear }\left(E_{*\_base}(t)\right)\right),
\end{equation}
where $\odot$ denotes the Hadamard product, $\sigma$ denotes sigmoid activation function, $*$ can be either $st$ or $ed$. The above gating process facilitates the model to pass only important information when constructing the graph.

\smallskip
\noindent\textbf{Learning Important Structure. }
Having obtained node embeddings $E_{st\_gate}(t)$ and $E_{ed\_gate}(t)$, we compute the weight of the adjacency matrix $\mathcal{A}_t$.
Let $\omega_{{i} {j}}^t$ denote the edge weight of adjacency matrix $\mathcal{A}_{t}$, where $v_{i}, v_{j} \in\mathcal{V} \text{ and } (v_{i},v_{j})$ $\in \mathcal{E}_{t}$. 
At each time step $t$, the node embeddings $E_{st\_gate}(t) = (e_{st, 1}^{t}, e_{st, 2}^{t},$ $\cdots, e_{st, N}^{t}) \in \mathbb{R}^{N \times d}$, $E_{ed\_gate}(t) = (e_{ed, 1}^{t}, e_{ed, 2}^{t}, \cdots, e_{ed, N}^{t})$ $\in \mathbb{R}^{N \times d}$ are used to construct the edge weight $\omega$ as follows:
\begin{equation}
\omega_{{i} {j}}^t=\textsf{Linear}\left(\operatorname{Tanh}\left(\left[e_{st, {i}}^{t} ; e_{ed, {j}}^{t}\right]\right)\right), \quad \text{ where } v_{i}, v_{j} \in \mathcal{V},
\end{equation}
where \textsf{Linear} is a linear layer without non-linearity. Note that $\omega_{ij}^t$ determines how likely $v_i$ and $v_j$ are connected at time step $t$. As learning progresses, the $\omega$ value continuously increases or decreases, which leads the $\omega$ value to be very large or very small. This causes the distribution of $\{\omega_{11}^t, \omega_{12}^t, \cdots, \omega_{nn}^t\}$ to have a high variance, and disturbs the construction of a valid dynamic graph at each time step $t$. To prevent this, we normalize $\{\omega_{11}^t, \omega_{12}^t, \cdots, \omega_{nn}^t\}$.
\begin{equation}
\omega_{11}^t, \omega_{12}^t, \cdots, \omega_{nn}^t \leftarrow \operatorname{normalize}\left(\omega_{11}^t, \omega_{12}^t, \cdots, \omega_{nn}^t\right),
\end{equation}
where $\operatorname{normalize}\left( \cdot \right)$ function normalizes the mean and standard deviation of the input data into 0 and $\alpha$, respectively. $\alpha $ is a hyperparameter that controls the influence of the neural network, and we set $\alpha = 1$. Finally, sigmoid is used as the activation function. 
Then, to construct a dynamic graph based on the structure that is important at time step $t$, we associate each edge $(v_{i},v_{j})\in\mathcal{E}_t$ with a random variable $p_{ij}^t\sim \text{Bernoulli}(\omega_{ij}^t)$, where $(v_{i},v_{j})$ remains in $\mathcal{E}_t$ if $p_{ij}^t=1$, and is dropped otherwise. Our goal is to make $p^t_{ij}$ learnable so that the constructed graph preserves the important structure at time step $t$.  However, as $p_{ij}^t=1$ is a discrete random variable, we cannot directly calculate the gradient of $ \omega_{ij}^t$, which hinders an end-to-end training of the model. Hence, we relax the discrete $p_{ij}^t=1$ to be a continuous variable in (0, 1) by using the Gumbel-Softmax reparameterization trick \cite{gumbel,gumbel2} as follows:
\begin{equation}
\small
p_{{i} {j}}^t \leftarrow \operatorname{Sigmoid}((\log {\delta} - \log(1-\delta) + \log{\omega_{{i} {j}}^t}- \log(1-\omega_{{i} {j}}^t))/\tau),
\end{equation}
where $\delta \sim \operatorname{Uniform}(0,1)$, $\tau$ is a temperature hyperarameter.
As temperature hyper-parameter $\tau$ approaches $0$, $p_{{i} {j}}^t$ gets closer to being binary. We set $\tau=1$ as we want the edge weight to vary according to its importance.

The Gumbel-Softmax reparameterization trick is a commonly chosen method for graph structure learning~\cite{gsl}. By adopting the Gumbel-Softmax trick,~\proposed~can learn important edges based on the edge weight $\omega_{{i} {j}}^t$ rather than always selecting the edge with highest $\omega_{{i} {j}}^t$. This enables our graph structure learner to flexibly control the characteristics (e.g., sparsity) of the final output graph in data-driven manner.

\smallskip
\noindent{\textbf{Edge Sampling Strategy. }}
{We find that rather than incorporating all the edges whose weight $p_{i j}^t > 0$, randomly selecting edges not only makes the training more stable, but also leads to further improvements on the model performance.} More precisely, the random edge sample strategy is defined as follows:
\begin{equation}
\small
\mathcal{A}^t_{i j} = \begin{cases}0, & \text { if }    \gamma < \rho, \\ p^t_{i j}, & \text { otherwise, }\end{cases}
\label{eqn:sampling}
\end{equation}
where $\rho \sim \operatorname{Uniform}(0,1)$ and $\gamma$ is a hyperarameter. The hyperparameter $\gamma$ controls the sampling probability, e.g., if $\gamma = 0.4$, $40\%$ of the edges are sampled and used. 
Note that the above sampling strategy can be considered as a form of data augmentation, often referred to as `graph structural perturbation' in the literature on graph self-supervised learning~\cite{zhu2021graph,zhu2020deep,thakoor2021large}. This strategy effectively introduces robustness to perturbation (i.e., noise) in the learned representations, as it exposes the model to various structural changes in the training data.

\smallskip
\noindent{\textbf{Sampling: Gumbel-Softmax vs Edge. }}
In our model, both the Gumbel-Softmax Trick and Edge Sampling are used, each with a distinct purpose. The Gumbel-Softmax Trick involves sampling to approximate gradients for discrete variables, essential for learning the importance of edges in the adjacency matrix. Conversely, Edge Sampling is used as data augmentation, which enhances the model's robustness to data variations by generating varied samples from the learned distribution. In summary, the two sampling strategies serve different but crucial roles in model learning and generalization.

\smallskip
\subsubsection{Structure Information}
\label{sec:si}

\label{sec:struct_info}
In this section, we define the Structure Information of the road networks that consists of the Target Space, which is a set of nodes that are likely to be connected by edges. 
More precisely, we utilize an adjacency matrix $\mathcal{A}^{sp}$, in which nodes that exist consecutively to each other are connected (Equation~\ref{eq1}). We assume that each node in $\mathcal{A}^{sp}$ has a self-loop, that is, $\mathcal{A}_{i i}^{sp}=1$ for $i \in\{1, \cdots, N\}$.

\begin{definition}
\label{def:SI}
\textbf{(Structure Information $\mathcal{S}^{k}\in \{0,1\}^{N \times N}$)} Let $d(v_{i}, v_{j})$ denote the shortest path distance between $v_{i} \in \mathcal{V}$ and $v_{j} \in \mathcal{V}$ in terms of $\mathcal{A}^{sp}$. 
For $v_{i} \in \mathcal{V}$, we define the Target Space $\mathcal{T}$ of $v_{i}$ over $\mathcal{S}^{k}$ as $\mathcal{T}(\mathcal{S}^{k},v_{i}) = \{v_{j} \in \mathcal{V} \mid d(v_{i}, v_{j}) \leq k\}$, where $k \in \mathbb{N}$. 
In other words, $\mathcal{T}(\mathcal{S}^{k},v_{i})$ denotes the set of nodes whose shortest path distance from $v_i$ is less than or equal to $k$.
Then, each $v_{i} \in \mathcal{V}$ connects all $v_{j} \in \mathcal{T}(\mathcal{S}^{k},v_{i})$ with an edge weight $\mathcal{S}^{k}_{{i} {j}}=1$. $\mathcal{S}^{k}$ is formally defined in an adjacency form as follows:
\begin{equation}
\mathcal{S}_{ij}^{k} = \begin{cases}1, & d(v_{i}, v_{j}) \leq k \\ 0 , & \text { otherwise. }\end{cases}
\end{equation}
\end{definition}

Some existing methods construct an adjacency matrix by connecting all nodes within a certain distance, utilizing a purely distance-based approach that relies on the absolute distances between sensors. This reliance may introduce bias as the graph's connectivity heavily fluctuates based on distance, potentially distorting the accurate representation of the sensor network and its inherent traffic patterns. In contrast, our proposed Structure Information ${S}^{k}$, considering all consecutive sensors within $k$-hops from a specific sensor, encapsulates local characteristics and maintains the relative spatial positioning among sensors. 
This comprehensive view is particularly valuable for capturing and analyzing spatial patterns and trends. Consequently, $\mathcal{S}^{k}$ offers a less biased, more nuanced perspective of the sensor network, effectively preserving relative spatial information.

\smallskip
\subsubsection{Adaptive Structure Information}
\label{sec:ASI}
To incorporate a macroscopic view of the road networks, the most straightforward and intuitive approach is to select from $\mathcal{A}^t$ only the nodes that belong to $\mathcal{S}^k$. More formally, the edge weight of the adjacency matrix $\mathcal{A}_{i j}^t$ at time step $t$ can be described as follows:

\begin{equation}
\label{eq:SI}
\mathcal{A}_{i j}^t= \begin{cases}0, & \text { if } v_j \notin \mathcal{T}(\mathcal{S}^{k},v_{i}), \\ p_{i j}^t, & \text { otherwise.}\end{cases}
\end{equation}

By integrating $\mathcal{S}^k$ and considering $k$ as a hyperparameter, this framework effectively encapsulates the macroscopic structural patterns within the road networks. 
However, rather than uniformly applying the same $k$-hop range for all nodes and all time steps, we propose a more adaptive model that considers varying hop ranges for each node and time step. Assume we have a Structure Information Group, denoted as $\mathcal{SG}_L = \{\mathcal{S}^1, \mathcal{S}^2, \ldots , \mathcal{S}^L\}$.

\begin{definition} 
\label{def:TRS} 
\textbf{(Hop Range Selector  \( R_t: \mathcal{V} \rightarrow \{1, 2, 3, \ldots,  L\} \))} 
The Hop Range Selector $R_t(v_i)$ determines the number of hops that a node $v_{i}$ should consider in the network $\mathcal{A}^{sp}$ at time $t$. This function assigns to each node a natural number $l \in \{1, 2, 3, \ldots, L\}$, representing the hop range that $v_i$ is deemed to interact with within the network at that time.
\end{definition}

\begin{definition}
\label{def:ASI}
\textbf{(Adaptive Structure Information \( \mathcal{I}_t  \in \{0,1\}^{N \times N} \))} 
For node \( v_i \in \mathcal{V}\) at time \( t \), we define the Dynamic Target Space  \( \mathcal{T}_\mathcal{D}  \) of \( v_i \)  over \( \mathcal{I}_{t}\), as the set of nodes that are within a certain hop range \( h \), which maximizes the probability of the Hop Range Selector \( R_t \). This is formally given by:
\begin{equation}
\mathcal{T}_\mathcal{D}(\mathcal{I}_{t},v_{i}) := \mathcal{T}(\mathcal{S}^{h},v_{i}),
\end{equation}
where \( h \) is chosen as the hop range for which \( v_i \) has the highest probability of needing to consider, based on \( R_t \), out of all possible hop range \( l \in \{1, 2, 3, \ldots, L\}\). Thus, \( h \) is determined by:
\begin{equation}
\label{eq:argmax}
h = \underset{l}{\operatorname{argmax}} \ P(R_t(v_i) = l).
\end{equation}

For the given time step \( t \), each node \( v_{i} \in \mathcal{V} \) is then connected to all nodes within its Dynamic Target Space \(  v_j \in \mathcal{T}_\mathcal{D}(\mathcal{I}_{t},v_{i}) \) with an edge weight \((\mathcal{I}_t)_{ij} = 1 \).  $\mathcal{I}_{t}$ is formally defined in an adjacency form as follows:
\begin{equation}
(\mathcal{I}_t)_{ij} = 
\begin{cases}
1, & \text{if } v_j \in \mathcal{T}_\mathcal{D}(\mathcal{I}_{t},v_{i}) \\
0, & \text{otherwise}.
\end{cases}
\end{equation}
\end{definition}

Now, the remaining issue is how to model $P(R_t(v_i) = l)$. To effectively model the nature of time series, we again leverage Recurrent Neural Networks (RNNs). Similar to the approach described in Section \ref{sec:DGCB_DNE}, we first define the initial node embedding $E_{h\_0} \in \mathbb{R}^{N \times m}$, where $N$ and $m$ denote the number of nodes and the dimension size of node embeddings, respectively. By using the Gated Recurrent Unit (GRU)~\cite{gru}, we obtain the intermediate dynamics of hidden states $E_{h}(t) \in \mathbb{R}^{N \times m}$ for nodes as follows:

\begin{equation}
E_{h}(t-1) = \textsf {GRU-Cell}(E_{h}(t), \mathbf{X}^{(t-1)}).
\end{equation}

Let $E_{h}(t)_i \in \mathbb{R}^{m}$ denote the hidden state embedding of node $v_i$ at time step $t$. We introduce a neural network, denoted as $NN(\cdot) : \mathbb{R}^{m} \rightarrow \mathbb{R}^{L}$, which is composed of multiple linear layers culminating in a softmax function. Then, the probability $P(R_t(v_i) = l)$ is given by:

\begin{equation}
P(R_t(v_i) = l) = NN(E_{h}(t)_i)[l],
\end{equation}
where $l\in\{1, 2, 3, \ldots, L\}$. Through this network, we are able to learn Adaptive Structure Information for each time step $t$. Due to the utilization of the $\operatorname{argmax}$ function in Equation~(\ref{eq:argmax}), we incorporate the Gumbel-Softmax technique to permit gradient propagation during the learning process.

In the final step, to capture the macroscopic view of the road networks using Adaptive Structure Information, the adjacency matrix $\mathcal{A}^t$ is pruned to include only nodes that are part of $\mathcal{I}_t$. Formally, the edge weights in the adjacency matrix at each timestep are defined as follows:

\begin{equation}
\label{eq:ASI}
\mathcal{A}_{i j}^t= \begin{cases}0, & \text { if } v_j \notin \mathcal{T}_\mathcal{D}(\mathcal{I}_{t},v_{i}) \\ p_{i j}^t, & \text { otherwise }\end{cases}
\end{equation}

\subsection{Spatio-Temporal Block}
\label{sec:STB}

\subsubsection{Spatial Processing Layer (SPL)}
\label{sec:STB_SPL}
In SPL, the encoded input signal (Section~\ref{sec:IL}), and the adjacency matrix constructed by Dynamic Graph Construction Block (Section~\ref{sec:DGCB}) are used to capture the spatial dependencies between the roads. We adopt diffusion convolution layer \cite{dcrnn}, which models the spatial dependency of traffic flow as a diffusion process. Through forward and reverse direction diffusion processes, diffusion convolution provides flexibility to models that use asymmetric adjacency matrices. 
More precisely, the diffusion convolution operation over the graph signal $\tilde{\mathbf{{X}}}\in \mathbb{R}^{N \times D}$, filter $f_{\theta}$, and the adjacency matrix $\mathcal{A}\in \mathbb{R}^{N \times N}$ is defined as follows:
\begin{equation}
\begin{aligned}
\tilde{\mathbf{X}}_{:, p} \star_{\mathcal{G}} f_{\boldsymbol{\theta}} = & \sum_{k=0}^{K-1}\left(\theta_{k, 1}\left(\boldsymbol{D}_{O}^{-1} \mathcal{A}\right)^{k}\right) \tilde{\mathbf{X}}_{:, p} \\
& + \sum_{k=0}^{K-1}\left(\theta_{k, 2}\left(\boldsymbol{D}_{I}^{-1} \mathcal{A}^{\top}\right)^{k}\right) \tilde{\mathbf{X}}_{:, p}
\end{aligned}
\end{equation}
where $p \in\{1, \cdots, D\}$, $\boldsymbol{\theta} \in \mathbb{R}^{K \times 2}$ is a learnable matrix, $K$ is the number of the diffusion steps. $\boldsymbol{D}_{O}^{-1} \mathcal{A}$ and $\boldsymbol{D}_{I}^{-1} \mathcal{A}^{\top}$ are state transition matrices. Here
$\boldsymbol{D}_{\boldsymbol{O}}=\operatorname{diag}(\mathcal{A} \mathbf{1})$ is the out-degree diagonal matrix, $\boldsymbol{D}_{\boldsymbol{I}}=\operatorname{diag}(\mathcal{A}^{\top} \mathbf{1})$ is the in-degree diagonal matrix, and $\mathbf{1} \in \mathbb{R}^{N}$ denotes all-one vector. 
With the convolution operation $\star_{\mathcal{G}}$, the Diffusion Convolutional Layer that maps $D$-dimensional features to $D'$-dimensional features can be defined as follows:
\begin{equation}
\small
\boldsymbol{H}_{:, q}=\sum_{p=1}^{D} {\tilde{\mathbf{X}}}_{:, p} \star_{\mathcal{G}} f_{\boldsymbol{\Theta}_{q, p,:,:}} \text { for } q \in\{1, \cdots, D'\},
\end{equation}
where ${\Theta} \in \mathbb{R}^{D' \times D \times K \times 2}=[\boldsymbol{\theta}]_{q, p}$ denote parameter tensor. 
Given $\mathcal{G}_{t}=\{\mathcal{V}, \mathcal{E}_{t},\mathcal{A}_{t}\}$ constructed in the Graph Construction Layer at each time step $t$, we apply the 
Diffusion Convolutional Layer to transform the input signal $\tilde{\mathbf{X}}^{\left(t\right)}\in \mathbb{R}^{N \times D}$ at time step $t$ to $\boldsymbol{H}^{\left(t\right)}\in \mathbb{R}^{N \times D'}$ as follows:
\begin{equation}
\small
\boldsymbol{H}_{:, q}^{\left(t\right)}=\sum_{p=1}^{D} {\tilde{\mathbf{X}}}_{:, p}^{\left(t\right)} \star_{\mathcal{G}_{t}} f_{\boldsymbol{\Theta}_{q, p,:,:}} 
\end{equation}
where $q \in\{1, \cdots, D'\}$.  
Parameter $\boldsymbol{\Theta}$ is shared across all time steps. Here, we set the dimensions of $D$ and $D'$ to have the same dimension to use the residual connection for the training stability. By including the residual learning method, the node representation after SPL can be written as follows:
\begin{equation}
\small
\operatorname{SPL}\left({\tilde{\mathbf{X}}}^{(t)}, \mathcal{G}_{t} \right)=\operatorname{ReLU}\left(\boldsymbol{H}^{(t)}+{\tilde{\mathbf{X}}}^{(t)}\right).
\end{equation}

\smallskip
\subsubsection{Temporal Processing Layer (TPL)}
\label{sec:STB_TPL}
To extract the temporal dynamics from the traffic flows, we use  1D temporal convolution~\cite{stgcn} followed by the Gate Tanh Unit (GTU)~\cite{gate} along the temporal dimension. Let $\boldsymbol{\mathcal { X }} \in \mathbb{R}^{N \times T^{\prime} \times D}$ be the input hidden representation of TPL, where $N$ is the number of nodes, $T^{\prime}$ is the input time steps, $D$ is the input channel dimension. Using the convolution kernel $\Lambda \in \mathbb{R}^{Ks \times D \times 2 D}$, input $\boldsymbol{\mathcal { X }}$ is transformed into $[U,V] \in \mathbb{R}^{N\times\left(T^{\prime}-Ks+1\right) \times\left(2 D\right)}$, where $Ks$ is temporal convolution kernel size and $U,V \in \mathbb{R}^{N\times(T^{\prime}-K s+1) \times D}$. Then, the temporal convolution using GTU is defined as follows:
\begin{equation}
\boldsymbol{\mathcal { X }} * _{\mathcal{T}} \Lambda  = Tanh(U) \odot \sigma(V),
\end{equation}
where $U$,$V$ are inputs of the gates in GTU, $\odot$ denotes the Hadamard product, and $Tanh$ and $\sigma$ denote Tanh and Sigmoid activation functions, respectively. We also adopt the residual structure in TPL, which can be written as follows:
\begin{equation}
\operatorname{TPL}\left(\boldsymbol{\mathcal { X }} \right)=\operatorname{LayerNorm}\left(\boldsymbol{\mathcal { X }} * _{\mathcal{T}} \Lambda+\boldsymbol{\mathcal { X }}^{:,T^{\prime}-Ks+1:,:}\right),
\end{equation}
where $\operatorname{TPL}\left(\boldsymbol{\mathcal { X }} \right)\in \mathbb{R}^{N\times\left(T^{\prime}-Ks+1\right) \times D}$.

\begin{table*}[t]
\caption{Performance of traffic flow forecasting on PeMS03, PeMS04, PeMS07 and PeMS08 datasets.}
\centering
\small
\begin{tabular}{clccclcccclcccclcccc}
\midrule
\multirow{2}{*}{Model} &  & \multicolumn{3}{c}{PeMS03}                       &  &  & \multicolumn{3}{c}{PeMS04}                       &  &  & \multicolumn{3}{c}{PeMS07}                      &  &  & \multicolumn{3}{c}{PeMS08}                      \\ \cline{3-5} \cline{8-10} \cline{13-15} \cline{18-20} 
                       &  & MAE            & RMSE           & MAPE           &  &  & MAE            & RMSE           & MAPE           &  &  & MAE            & RMSE           & MAPE          &  &  & MAE            & RMSE           & MAPE          \\ \midrule
HA                     &  & 31.58          & 52.39          & 33.78          &  &  & 38.03          & 59.24          & 27.88          &  &  & 45.12          & 65.64          & 24.51         &  &  & 34.86          & 52.04          & 27.07         \\
ARIMA                  &  & 35.41          & 47.59          & 33.78          &  &  & 33.73          & 48.80          & 24.18          &  &  & 38.17          & 59.27          & 19.46         &  &  & 31.09          & 44.32          & 22.73         \\
FC-LSTM                &  & 21.33          & 35.11          & 23.33          &  &  & 26.24          & 40.49          & 19.30          &  &  & 29.96          & 43.94          & 14.34         &  &  & 22.20          & 33.06          & 15.02         \\
DCRNN                  &  & 18.18          & 30.31          & 18.91          &  &  & 24.70          & 38.12          & 17.12          &  &  & 25.30          & 38.58          & 11.66         &  &  & 17.86          & 27.83          & 11.45         \\
STGCN                  &  & 17.49          & 30.12          & 17.15          &  &  & 22.70          & 35.55          & 14.59          &  &  & 25.38          & 38.78          & 11.08         &  &  & 18.02          & 27.83          & 11.40         \\
ASTGCN                 &  & 17.69          & 29.66          & 19.40          &  &  & 22.93          & 35.22          & 16.56          &  &  & 28.05          & 42.57          & 13.92         &  &  & 18.61          & 28.16          & 13.08         \\
STSGCN                 &  & 17.48          & 29.21          & 16.78          &  &  & 21.19          & 33.65          & 13.90          &  &  & 24.26          & 39.03          & 10.21         &  &  & 17.13          & 26.80          & 10.96         \\
STFGNN                 &  & 16.77          & 28.34          & 16.30          &  &  & 19.83          & 31.88          & 13.02          &  &  & 22.07          & 35.80          & 9.21          &  &  & 16.64          & 26.22          & 10.60         \\
AGCRN                  &  & 15.98          & 28.25          & 15.23          &  &  & 19.83          & 32.26          & 12.97          &  &  & 22.37          & 36.55          & 9.12          &  &  & 15.95          & 25.22          & 10.09         \\
STGODE                 &  & 16.50          & 27.84          & 16.69          &  &  & 20.84          & 32.82          & 13.77          &  &  & 22.99          & 37.54          & 10.14         &  &  & 16.81          & 25.97          & 10.62         \\
Z-GCNETs               &  & 16.64          & 28.15          & 16.39          &  &  & 19.50          & 31.61          & 12.78          &  &  & 21.77          & 35.17          & 9.25          &  &  & 15.76          & 25.11          & 10.01         \\
DSTAGNN                &  & {\ul 15.57}    & {\ul 27.21}    & {\ul 14.68}    &  &  & {\ul 19.30}    & {\ul 31.46}    & {\ul 12.70}    &  &  & {\ul 21.42}    & {\ul 34.51}    & {\ul 9.01}    &  &  & {\ul 15.67}    & {\ul 24.77}    & {\ul 9.94}    \\ \midrule
\proposed                  &  & \textbf{15.15} & \textbf{26.49} & \textbf{14.47} &  &  & \textbf{18.84} & \textbf{31.03} & \textbf{12.39} &  &  & \textbf{20.58} & \textbf{34.49} & \textbf{8.66} &  &  & \textbf{15.28} & \textbf{24.58} & \textbf{9.78} \\ \midrule
\end{tabular}
\label{tab:result}
\end{table*}


\begin{table}[t]
\caption{Datasets description.}
\small
\centering
\begin{tabular}{c|cccc}
\midrule
Dataset & Nodes & Edges & Time step & Time range            \\ \midrule
PeMS03  & 358   & 547  & 26,208     & 9/1/2018 - 11/30/2018 \\
PeMS04  & 307   & 340  & 16,992     & 1/1/2018 - 2/28/2018  \\
PeMS07  & 883   & 866  & 28,224     & 5/1/2017 - 8/31/2017  \\
PeMS08  & 170   & 295  & 17,856     & 7/1/2016 - 8/31/2016  \\ \midrule
\end{tabular}
\label{tab:data}
\end{table}
\smallskip

\smallskip
\subsubsection{Output Layer}
Since padding is not involved when performing 1D temporal convolution in TPL, the time dimension of the data is compressed as it is processed by several consecutive Spatio-Temporal Layers. In the output layer, to compress all the temporal dimension information, 1D temporal convolution is performed using a temporal kernel with the same size as the temporal dimension of the input hidden representation. Therefore, the result of the output layer has $\mathbb{R}^{N \times D}$ dimension.

\smallskip
\subsection{Prediction Layer}
\label{sec:PL}
As shown in Figure~\ref{fig3},~\proposed~consists of multiple Spatio-Temporal Blocks. Let $\boldsymbol{\mathcal {O}_{i}} \in \mathbb{R}^{N \times D}$ be the output of the $i$-th Spatio-Temporal Block. Then, we concatenate all outputs of Spatial-Temporal Blocks along the feature dimension, and feed it into the Prediction Layer, which is composed of a Linear Layer and produces predictions for future $T$ time steps as follows:
\begin{equation}
\left[\hat{\mathbf{X}}^{\left(t+1\right)}, \cdots , \hat{\mathbf{X}}^{\left(t+T\right)} \right]= \left[\boldsymbol{\mathcal { O }}_{1}, \cdots, \boldsymbol{\mathcal { O }}_{n}\right] W_{p}+b_{p},
\end{equation}
where $n$ is the number of Spatio-Temporal Blocks, $W_{p} \in \mathbb{R}^{nD \times T \times F}$ is a learnable parameter and $b_{p}\in \mathbb{R}^{T \times F}$ is the bias, which is added node-wise. Here $F$ is the number of features we want to predict. 
We set $F = 1$ because we want to predict the traffic flow from the model.
\smallskip
\subsection{Loss Function}
\label{sec:LF}
Let $\left[\mathbf{X}^{\left(t-T^{\prime}+1\right)}, \cdots , \mathbf{X}^{t} \right] $ be the input for~\proposed. Then, the model predicts values $\left[\hat{\mathbf{X}}^{\left(t+1\right)}, \cdots , \hat{\mathbf{X}}^{\left(t+T\right)} \right]$ for the future $T$ time steps. We use mean absolute error (MAE) as the loss function for training the model as follows:
\begin{equation}
Loss=\frac{\sum_{i=1}^{i=T} \sum_{j=1}^{j=N} \sum_{k=1}^{k=F}\left|\hat{\mathbf{X}}_{j k}^{(t+i)}-\mathbf{X}_{j k}^{(t+i)}\right|}{T \times N \times F}.
\end{equation}

\smallskip
\section{Experiments}
\smallskip
\subsection{Experimental Setup}
\noindent{\textbf{Datasets}. }
In order to evaluate the performance of the~\proposed, we conduct experiments on four real-world highway traffic datasets, i.e., PeMS03, PeMS04, PeMS07 and PeMS08, which are collected by the Caltrans Performance Measurement System (PeMS) in real time every 30 seconds~\cite{pems}. The traffic data are aggregated into 5-minutes intervals, which means there are 12 time steps in the traffic flow per hour (i.e., $T=12$). The input data is re-scaled using Z-score normalization. The details of the datasets are summarized in Table~\ref{tab:data}.

\smallskip
\noindent{\textbf{Baselines. }}
We compare~\proposed~with the following baselines:
\begin{itemize}[leftmargin=5mm]
\item \textbf{HA}~\cite{hamilton2020time} uses an average value of inputs as prediction of the next time step value.
\item \textbf{ARIMA}~\cite{arima} is a representative classical time series model for predicting future values.
\item \textbf{FC-LSTM}~\cite{fclstm} uses RNNs with fully connected LSTM hidden units.
\item \textbf{DCRNN}~\cite{dcrnn} combines diffusion convolution with recurrent neural networks in a sequence-to-sequence manner.
\item \textbf{STGCN}~\cite{stgcn} combines graph convolutional structure with gated temporal convolution. 
\item \textbf{ASTGCN}~\cite{astgcn} utilizes spatial and temporal attention mechanisms with spatial-temporal convolution to learn the dynamic spatial-temporal correlations.  
\item \textbf{STSGCN}~\cite{stsgcn} utilizes localized spatial-temporal graph convolutional module to synchronously capture the complex localized spatial-temporal correlations directly.
\item \textbf{STFGNN}~\cite{stfgnn} captures local and global correlations simultaneously, by assembling a Gated dilated CNN module with spatial-temporal fusion graph module.
\item \textbf{AGCRN}~\cite{agcrn} utilizes data adaptive graph generation module to infer the inter-dependencies among different traffic series.
\item \textbf{STGODE}~\cite{stgode} extracts longer-range spatial-temporal correlations using continuous representation of GNNs.
\item \textbf{Z-GCNETs}~\cite{zgcnnet} utilizes time-aware zigzag persistence and zigzag topological layer for time-aware graph convolutional networks.
\item \textbf{DSTAGNN}~\cite{dstagnn} utilizes spatial-temporal attention module and gated convolution module to exploit dynamic spatial-temporal correlation.
\end{itemize}

\begin{figure}[t]
\centerline{\includegraphics[width=0.95\linewidth]{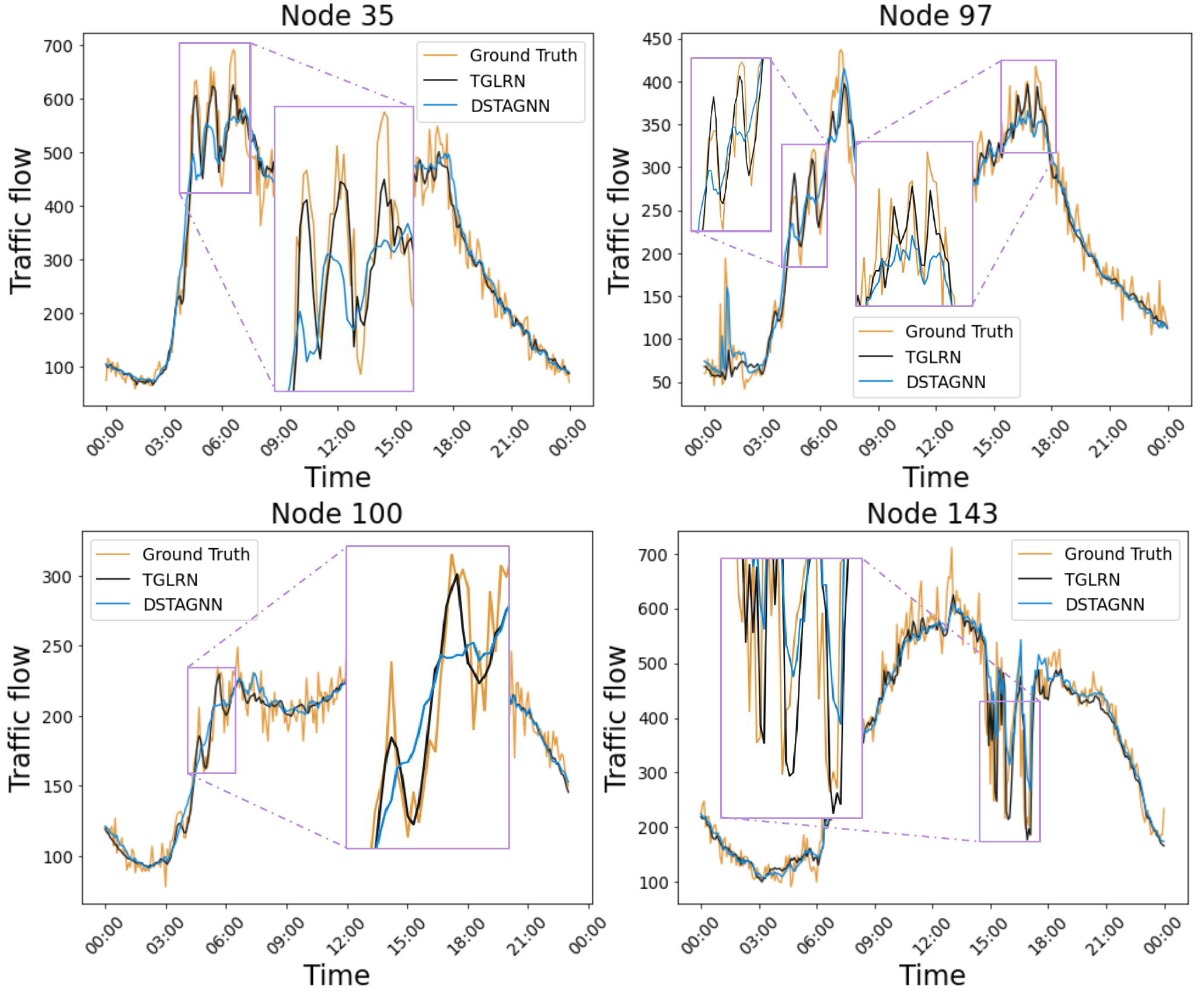}}
\caption{Prediction curves of DSTAGNN and~\proposed~on PeMS08 dataset.}
\label{plot1}
\vspace{-1ex}
\end{figure}

\begin{figure}[t]
\centerline{\includegraphics[width=0.95\linewidth]{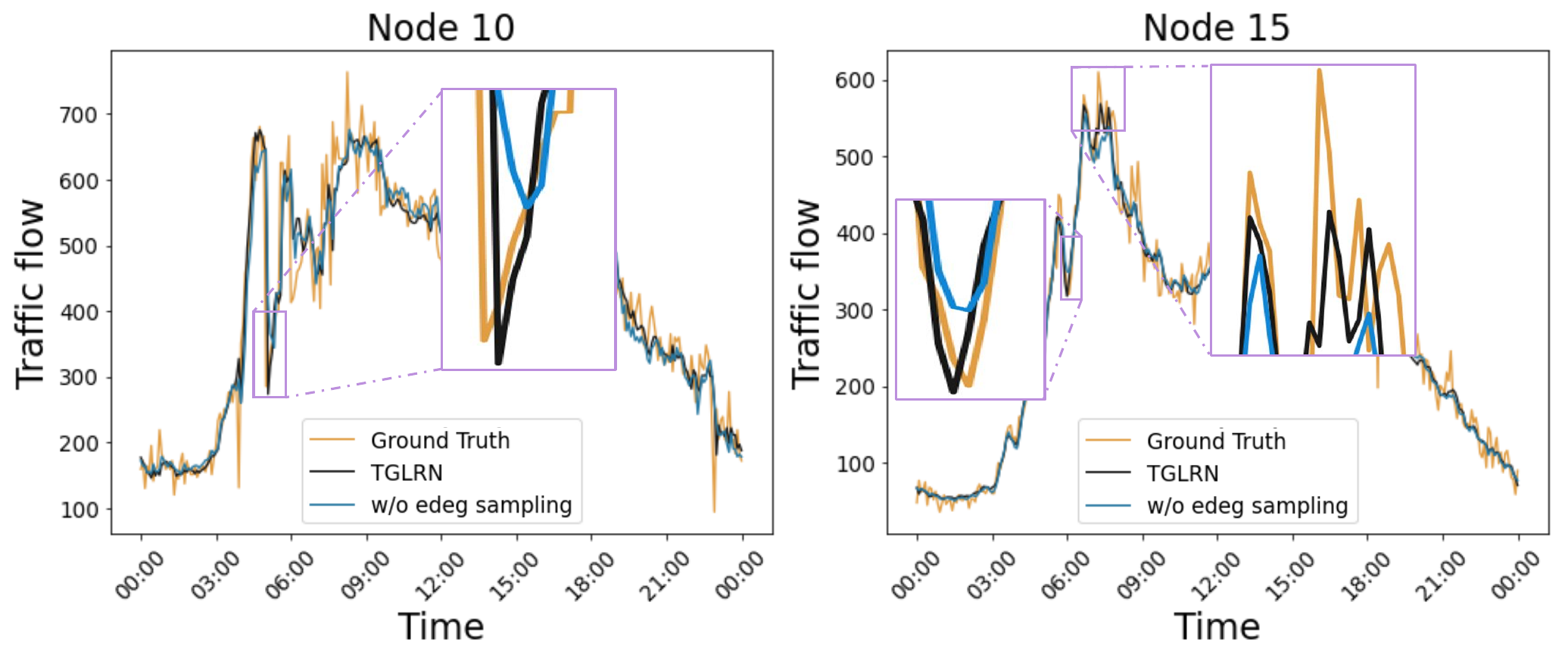}}
\caption{Prediction curves of~\proposed~and~\proposed~w/o edge sampling on PeMS08 dataset.}
\label{plot1_1}
\vspace{-1ex}
\end{figure}

\smallskip                    
\noindent{\textbf{Evaluation Protocol.} }
We adopt three commonly used evaluation metrics to evaluate the model performance: mean absolute error (MAE), root mean squared error (RMSE), and mean absolute percentage error (MAPE).
All datasets are split with a ratio of 6:2:2 into training sets, validation sets, and test sets. We use one hour of historical data to predict the next 60 minutes data, that is, as datasets are aggregated into 5-minute windows, 12 time steps of historical data are used to predict future 12 time steps data.
Experiments are conducted on Intel(R) Xeon(R) Gold 6226R CPU @ 2.90GHz and NVIDIA RTX A6000. 

\smallskip
\noindent{\textbf{Implementation Details. }}
The detailed hyperparameter setting of~\proposed~is as follows: 
In Graph Construction Layer described in Section~\ref{sec:gcl}, the dimensionality of node embeddings (i.e., $d$) is tuned in $\{4,8,16,32,64\}$ and the edge sampling parameter $\gamma$ is tuned in $\{0.05,0.1,0.2,0.3\}$. As described in Section~\ref{sec:ASI}, we use the Structure Information Group for constructing graphs. Specifically, the Structure Information Group is tuned in $\{\mathcal{SG}_{5}, \mathcal{SG}_{7},\mathcal{SG}_{10},\mathcal{SG}_{15}\}$. We used two SPLs and TPLs in Spatio-Temporal Block, and the hidden dimension size is fixed to 64 (i.e., $
D=D'=64)$. The number of Spatio-Temporal Blocks in the model is tuned in $\{3,4,5,6\}$, the number of the diffusion steps (i.e., $K$) is fixed to 2 in SPL, the temporal convolution kernel size (i.e., $Ks$) is fixed to 6 in TPL. Learning rate is fixed to 0.005, and dropout rate is tuned in $\{0.05,0.1,0.15,0.2\}$.

\smallskip
\subsection{Experimental Results}
\smallskip
\noindent{\textbf{Overall Evaluation.}}
Table~\ref{tab:result} shows the results of traffic flow forecasting. We have the following observations:
\textbf{1)} Deep learning models outperform traditional statistical and machine learning-based time series prediction methods. 
\textbf{2)} Deep learning models using graph structures show better performance than FC-LSTM\cite{fclstm}, which does not consider the spatial dependencies between roads. \textbf{3)}~\proposed~outperforms other baseline models implying that dynamically learning a graph at each time step in the microscopic view, together with capturing the macroscopic view is beneficial.

\begin{figure}[t]
\centerline{\includegraphics[width=1\linewidth]{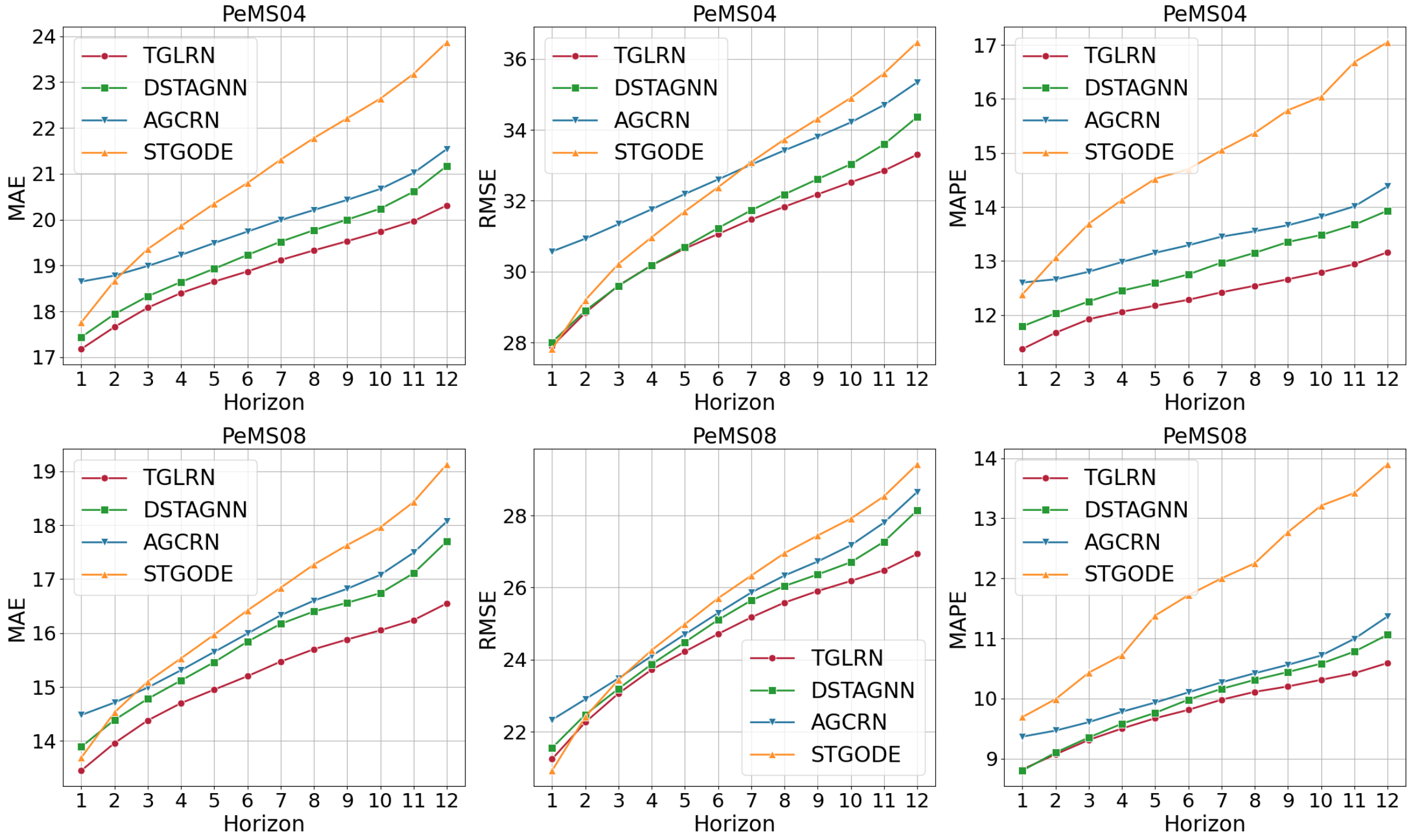}}
\caption{Prediction error at each horizon on PeMS04 and PeMS08 datasets.}
\label{plot2}
\vspace{-1ex}
\end{figure}

\smallskip
\noindent{\textbf{Traffic Forecasting Visualization. }}
Figure~\ref{plot1} depicts the ground truth, and the prediction curves of~\proposed~and DSTAGNN~\cite{dstagnn}, which is the state-of-the-art traffic forecasting method. 
We observe that \proposed~is superior to DSTAGNN in general, which verifies the benefit of our framework. Interestingly, we observe that~\proposed~outperforms DSTAGNN especially in cases where there are sudden changes of traffic flow (e.g., Node 100 and 143). 
{We show in Figure~\ref{plot1_1} that the edge sampling strategy further improves the accuracy of the traffic flow prediction.}

\smallskip
\noindent{\textbf{Prediction Performance over Horizon.}}
Figure~\ref{plot2} illustrates the prediction performance at each horizon (i.e., from $(t+1)$ to $(t+T)$) on PeMS04 and PeMS08 datasets. In general, as the predictive horizon increases, the slope of the performance lines increases, implying that the models perform worse in predicting distant future.
On the other hand, we observe that the increase in slope of the performance line of~\proposed~is relatively moderate compared with state-of-the-art baselines, which indicates the superiority of~\proposed~in predicting traffic flow of distant future.

\begin{figure}[ht]
\centerline{\includegraphics[width=1\linewidth]{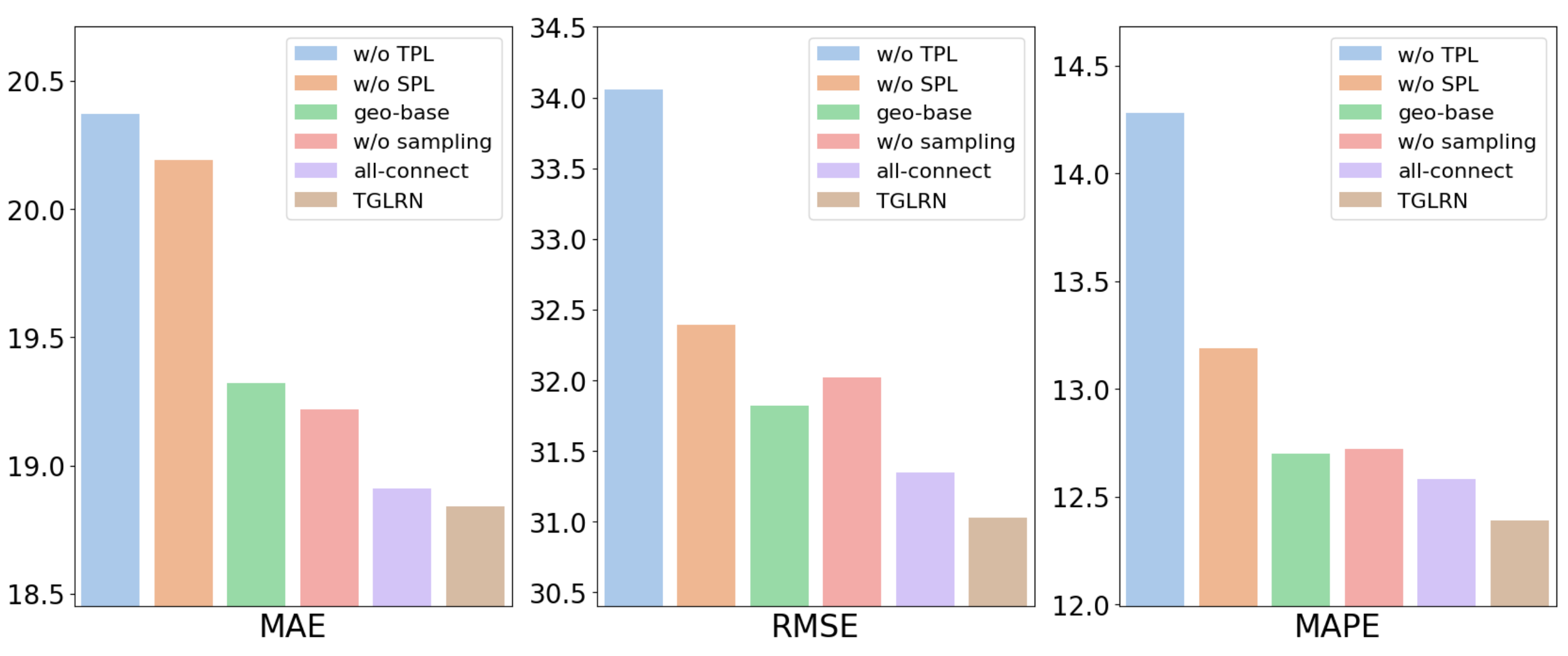}}
\caption{Component analysis of~\proposed.}
\label{bar1}
\vspace{-2ex}
\end{figure}

\smallskip
\subsection{Model Analysis}
\smallskip
\subsubsection{Component Analysis}
To investigate the effect of each component of~\proposed, we conduct additional experiments on PeMS04 dataset with five variants of~\proposed~described as follows:

\begin{itemize}[leftmargin=5mm]

\item w/o TPL: This model does not equip with TPL described in Section~\ref{sec:STB_TPL}.

\item w/o SPL: This model does not equip with SPL described in Section~\ref{sec:STB_SPL}.

\item geo-base: This model uses Structure
Information, denoted as \(\mathcal{S}^1\), by applying Equation (\ref{eq:SI}) instead of using Adaptive Structure Information, to consider only geographically connected nodes.

\item w/o sampling: This model does not adopt edge sampling strategy described in Section~\ref{sec:gcl}.

\item all-connect: This model uses a fully connected graph as the structure information ignoring the relative geographical distance between nodes.

\end{itemize}

\noindent We have the following observations in Figure~\ref{bar1}.
\textbf{1)} SPL and TPL significantly affect the prediction performance, and they are essential model components. Notably, TPL contributes more to the performance improvement than SPL, which implies that when forecasting based on historical temporal data, modeling temporal dependencies is more important than modeling spatial dependence.
\textbf{2)} The inferior results of geo-base and all-connect demonstrate the importance of using Adaptive Structure Information to appropriately restrict the geographic distances between node pairs in the graphs.
\textbf{3)} The results of w/o sampling demonstrate that our proposed edge sampling strategy is crucial and contributes to improving the model performance.

\begin{table}[htbp]
\caption{
Experiments on various structure information. Note that $\mathcal{S}^{k}$ is the integration of Structure Information, as defined in Definition~\ref{def:SI}, using Equation~(\ref{eq:SI}) across all time steps. Similarly, $\mathcal{I}$ represents the integration of Adaptive Structure Information defined in Definition~\ref{def:ASI} at each time step utilizing Equation~(\ref{eq:ASI}). Note that $\mathcal{S}^{\infty}$ indicates ~\proposed~ employing a fully connected graph as structure information, and $\mathcal{I}^{*}$ indicates ~\proposed~ that does not use a learned dynamic graph $\mathcal{A}^t$ at each time step (i.e., instead of applying Equation~(\ref{eq:ASI}), directly using adjacency matrices $\mathcal{I}_t$ composed of 0 or 1 for each time step).
}
\small
\centering
\begin{tabular}{c|c|ccc}
\midrule
\hspace{1ex}Dataset\hspace{1ex}                 & \hspace{1ex} Structure Information\hspace{1ex}   & \hspace{1ex} MAE\hspace{1ex}             & RMSE           & \hspace{1ex} MAPE\hspace{1ex}            \\ \midrule
\multirow{5}{*}{PeMS04} & $\mathcal{S}^{1}$                        & 19.32          & 31.82          & 12.70          \\
                        & $\mathcal{S}^{30}$                       & 18.94          & 31.32          & 12.55          \\
                        & $\mathcal{S}^{\infty}$                       & 18.90 & 31.33 & 12.56 \\ 
                        & $\mathcal{I}^{*}$                       & 18.92 & 31.46 & 12.49 \\
                        & $\mathcal{I}$                       & \textbf{18.84} & \textbf{31.03} & \textbf{12.39} \\
                        \midrule
\multirow{5}{*}{PeMS07} & $\mathcal{S}^{1}$                        & 21.39          & 35.45          & 8.95           \\
                        & $\mathcal{S}^{30}$                       & 20.99          & 34.69          & 8.75           \\
                        & $\mathcal{S}^{\infty}$                       & 20.98 & 34.93 & 8.93  \\
                        & $\mathcal{I}^{*}$                       & 20.76 & 34.78 & 8.68  \\
                        & $\mathcal{I}$                       & \textbf{20.58} & \textbf{34.49} & \textbf{8.66} \\
                        \midrule
\end{tabular}
\label{tab:structure}
\end{table}

\begin{table}[htbp]
\caption{Analysis on edge sampling strategy.}
\setlength{\extrarowheight}{2pt} 
\fontsize{7.7pt}{9}\selectfont
\centering
\begin{tabular}{c|c|c|ccc}
\midrule
\hspace{1ex}Dataset\hspace{1ex}                 & Noise                   & Edge Sampling?          & \hspace{1ex}MAE\hspace{1ex}   & RMSE  & \hspace{1ex}MAPE\hspace{1ex}  \\ \midrule
\multirow{4}{*}{PeMS03} & \multirow{2}{*}{$\mathcal{N}(0,5^2)$} & \xmark   & 15.63 & 27.47 & 15.05 \\
                        &                                       & \cmark   & 15.21 & 26.54 & 14.50 \\ \cline{2-6} 
                        & \multirow{2}{*}{$\mathcal{N}(0,10^2)$} & \xmark   & 15.72 & 27.54 & 15.19 \\
                        &                                       & \cmark   & 15.43 & 26.73 & 14.73 \\ \midrule
\multirow{4}{*}{PeMS04} & \multirow{2}{*}{$\mathcal{N}(0,5^2)$} & \xmark   & 19.28 & 32.07 & 12.83 \\
                        &                                       & \cmark   & 18.94 & 31.08 & 12.52 \\ \cline{2-6} 
                        & \multirow{2}{*}{$\mathcal{N}(0,10^2)$} & \xmark   & 19.43 & 32.15 & 13.12 \\
                        &                                       & \cmark   & 19.20 & 31.25 & 12.90 \\ \midrule
\end{tabular}
\label{tab:edgesampling}
\end{table}

\smallskip
\subsubsection{Analysis on Structure Information} In Table~\ref{tab:structure}, we have the following observations: 
\textbf{1)} 
The model performance degrades when constructing a graph using roads that are too distant (i.e., $\mathcal{S}^{30}$ and $\mathcal{S}^{\infty}$). 
We argue that in the macroscopic view, as roads in geographically distant areas have different trends (Figure~\ref{fig2}), considering too distant roads is not helpful for predicting the traffic flow of each road.
\textbf{2)} 
The model performance degrades when constructing a graph using roads that are too close (i.e., $\mathcal{S}^{1}$),  whereas the performance is optimal when applying our proposed Adaptive Structure Information (i.e., $\mathcal{I}$). These results underscore the effectiveness of our Adaptive Structure Information method by optimally selecting a search range for each node at every time step.
\textbf{3)} Considering dynamic spatial dependencies over time consistently outperforms the case when the model is trained solely based on a static graph (i.e., $\mathcal{I}^{*}$). We argue that in the microscopic view, as each time step inherently carries different spatial dependencies, considering spatial dependencies between roads that change over time is crucial.

\subsubsection{Analysis on edge sampling strategy.}
We conduct an experiment to confirm whether the edge sampling method affects model's robustness. We introduce zero mean Gaussian noise $\mathcal{N}(0,\sigma^{2})$ into input data when testing our model. In Table ~\ref{tab:edgesampling}, we observe that applying the edge sampling strategy consistently superior to the case when the edge sampling strategy is not applied, which means structural perturbation actually gives robustness to the representations.

\begin{figure}[t]
\centerline{\includegraphics[width=1\linewidth]{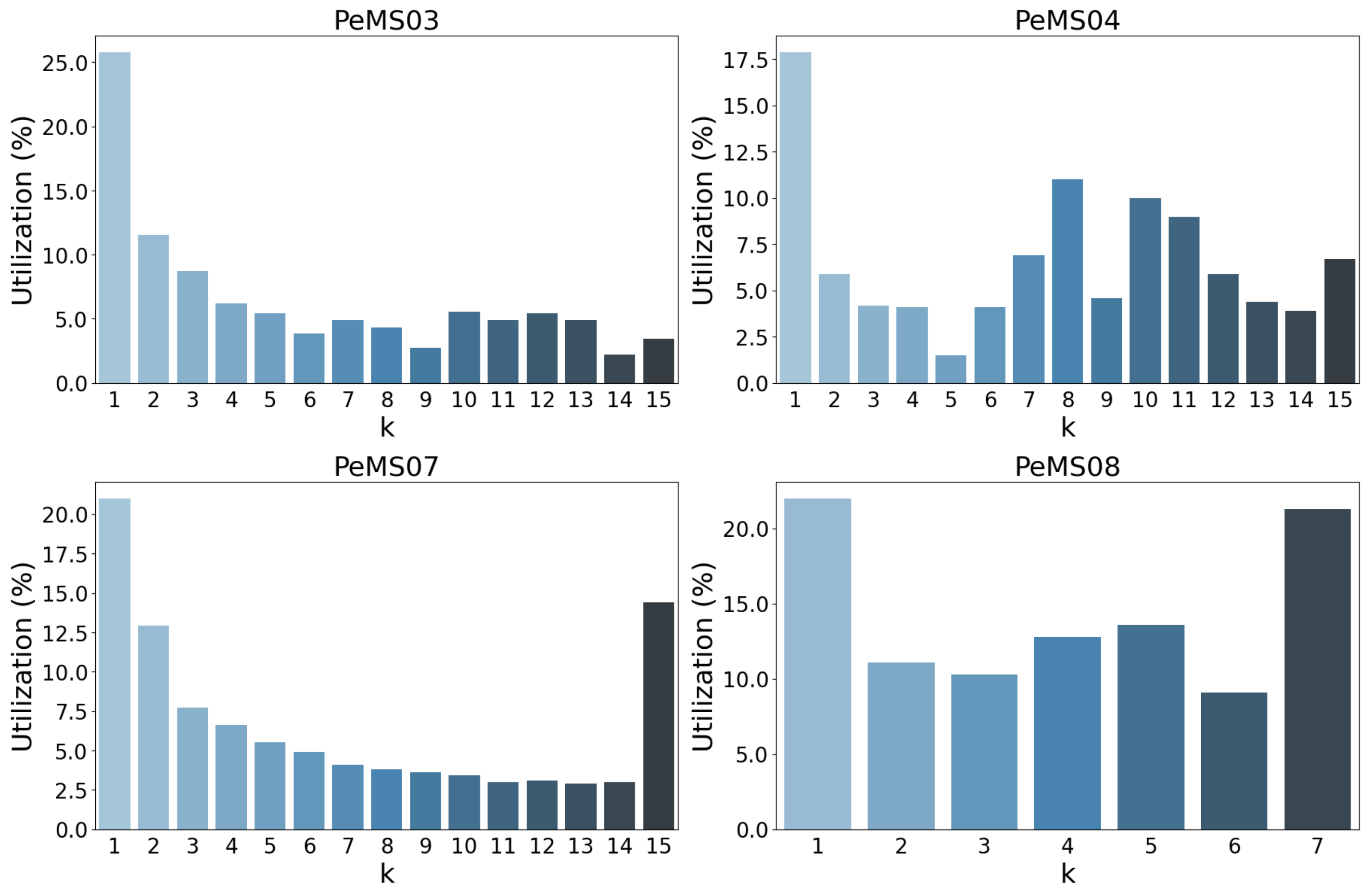}}
\caption{Visualization of the utilization of each Structure Information $\mathcal{S}^k$. The x-axis represents the Structure Information $\mathcal{S}^k$, and the y-axis shows the proportion of Structure Information selected via the Hop Range Selector.}
\label{fig:ASI_bar}
\end{figure}

\begin{table}[t]
\caption{Analysis on Gumbel-Softmax trick.}
\small
\centering
\begin{tabular}{c|c|ccc}
\midrule
\hspace{1ex}Dataset\hspace{1ex}                   & Gumbel-Softmax Trick?   & \hspace{1ex}MAE\hspace{1ex}     & \hspace{1ex}RMSE\hspace{1ex}    & \hspace{1ex}MAPE\hspace{1ex}  \\ \midrule
\multirow{2}{*}{PeMS03}   & \xmark      & 15.24   & 26.57   & 14.77 \\
                          & \cmark          & 15.15   & 26.49   & 14.47 \\ \midrule
\multirow{2}{*}{PeMS08}   & \xmark     & 15.33   & 24.70   & 9.84  \\
                          & \cmark          & 15.28   & 24.58   & 9.78  \\ \midrule
\end{tabular}
\label{tab:gumbel}
\end{table}

\smallskip
\subsubsection{Analysis on Gumbel-Softmax Trick} 
We conduct experiments to validate the benefit of training~\proposed~based on important structure learned by the Gumbel-Softmax Trick. In Table~\ref{tab:gumbel}, we observe that we achieve further improvements in the overall performance when using the Gumbel-Softmax trick rather than directly using sigmoid on edge weights. We attribute this to the fact that Gumbel-Softmax trick samples meaningful nodes by capturing the importance of edges. However, it is important to note that even without the Gumbel-Softmax trick, we still achieve the state-of-the-art performance.

\smallskip
\subsubsection{ Visualization of Hop Range Selection Using Adaptive Structure Information}
In Section~\ref{sec:ASI}, we defined Adaptive Structure Information, enabling the model to learn suitable hop ranges. The rationale for this approach, as discussed in Section~\ref{sec:PD}, is that each dataset has different criteria and methods for constructing adjacency matrices $\mathcal{A}^{s p}$ based on spatial neighbor relationships. Additionally, the optimal hop range might vary for each node and each time step. Through Figure~\ref{fig:ASI_bar}, we visualized  utilization of each Structure Information $ \mathcal{S}^k $ within the Structure Information Group $\mathcal{SG}_L = \{\mathcal{S}^1, \mathcal{S}^2, \ldots , \mathcal{S}^L\}$ for various datasets. These results demonstrate that the distribution of Structure Information utilization, selected via the Hop Range Selector, differs across datasets, indicating that the range of consideration varies for each node and time step for optimal performance. Notably, for dataset PeMS07, the high utilization of $S^{15}$ was observed. In such cases, we conducted experiments to explore a broader range by setting the hyperparameter $L$ to values greater than 15, but these experiments were observed not to have a significant impact on performance improvement.  This suggests that beyond a certain range, the additional Structure Information may not yield meaningful improvements, potentially leading to redundant information.

\begin{table}[htbp]
\caption{Training time spent per epoch.}
\setlength{\extrarowheight}{2pt} 
\fontsize{8.5pt}{9.5}\selectfont
\centering
    \begin{tabular}{ccc|ccccccccl}
    \midrule
     & \multirow{2}{*}{Dataset} &  & \multicolumn{3}{c}{Training time per epoch (sec.)}                                   \\ \cline{4-6} 
     &                          &  & \hspace{2ex}\proposed\hspace{2ex} & DSTAGNN & \hspace{2ex}Z-GCNETs\hspace{2ex} \\ \midrule
     & PeMS04                   &  & 72.61 & 113.24  & 39.66   \\ \midrule
     & PeMS08                   &  & 57.60 & 108.62  & 34.85   \\ \midrule
    \end{tabular}
\label{tab:time_epoch}
\end{table}

\begin{table}[t]
\caption{Number of model parameters.}
\setlength{\extrarowheight}{2pt} 
\fontsize{8.5pt}{9.5}\selectfont
\centering
\begin{tabular}{ccc|ccccccccl}
\midrule
 & \multirow{2}{*}{Dataset} &  & \multicolumn{3}{c}{Number of model parameters}                                           \\ \cline{4-6} 
 &                          &  & \hspace{2ex}\proposed\hspace{2ex}   & DSTAGNN & \hspace{2ex}Z-GCNETs\hspace{2ex} \\ \midrule
 & PeMS04                   &  &  2,087,766   & 3,579,728 &         455,034             \\ \midrule
 & PeMS08                   &  & 2,083,631 & 2,296,860 & 453,664   \\ \midrule
\end{tabular}
\label{tab:space_complexity}
\end{table}

\smallskip
\subsection{Efficiency Analysis}
\smallskip
\subsubsection{Training Time Comparison}
In Table~\ref{tab:time_epoch}, we compare the training time spent per epoch for training~\proposed, DSTAGNN~\cite{dstagnn}, and Z-GCNETs~\cite{zgcnnet}. We observed  that \proposed~spent a moderate amount of time per epoch despite its advanced architecture design, which demonstrates that~\proposed~is scalable.

\smallskip
\subsubsection{Space Complexity Analysis}
In Table~\ref{tab:space_complexity}, we compare the number of model parameters of~\proposed~with two recent state-of-the-art methods, i.e., DSTAGNN~\cite{dstagnn} and Z-GCNETs~\cite{zgcnnet}. Specifically, \proposed~outperforms DSTAGNN even with less number of parameters. Moreover, although Z-GCNETs requires a relatively small number of model parameters, it is greatly outperformed by~\proposed.
Note that the Graph Construction Layer requires $O\left(T^{\prime}N^{2}\right)$ space complexity, which means~\proposed~consumes more memory as the number of nodes $N$ increases. 
This becomes a problem when sensors are expanded, or new sensors are installed in new areas. One way to solve this problem is dividing the area based on a geographical location and training separate models, which we leave as future work.

\section{Conclusion}
In this paper, we propose a novel neural network model, called Temporal Graph Learning Recurrent Neural Network (\proposed), for traffic flow forecasting.~\proposed~effectively generates and trains graphs with different semantic information according to time changes through the RNN structure (i.e., microscopic view). 
In the process of constructing graphs, we train the model using Adaptive Structure Information to select an optimal range for connecting node pairs in the graphs, aiming to exclude unnecessary information during the training (i.e., macroscopic view). In addition, we devise an edge sampling strategy when constructing the graph at each time step to endow the model with robustness, which eventually leads to further improvements. Experiments on four real-world datasets demonstrate that~\proposed~captures the complex spatio-temporal dependencies of road networks, and provides superior performance of traffic flow forecasting.

\bibliographystyle{ACM-Reference-Format}
\bibliography{sample-base}


\begin{thebibliography}{47}


\ifx \showCODEN    \undefined \def \showCODEN     #1{\unskip}     \fi
\ifx \showDOI      \undefined \def \showDOI       #1{#1}\fi
\ifx \showISBNx    \undefined \def \showISBNx     #1{\unskip}     \fi
\ifx \showISBNxiii \undefined \def \showISBNxiii  #1{\unskip}     \fi
\ifx \showISSN     \undefined \def \showISSN      #1{\unskip}     \fi
\ifx \showLCCN     \undefined \def \showLCCN      #1{\unskip}     \fi
\ifx \shownote     \undefined \def \shownote      #1{#1}          \fi
\ifx \showarticletitle \undefined \def \showarticletitle #1{#1}   \fi
\ifx \showURL      \undefined \def \showURL       {\relax}        \fi
\providecommand\bibfield[2]{#2}
\providecommand\bibinfo[2]{#2}
\providecommand\natexlab[1]{#1}
\providecommand\showeprint[2][]{arXiv:#2}

\bibitem[Bai et~al\mbox{.}(2020)]%
        {agcrn}
\bibfield{author}{\bibinfo{person}{Lei Bai}, \bibinfo{person}{Lina Yao}, \bibinfo{person}{Can Li}, \bibinfo{person}{Xianzhi Wang}, {and} \bibinfo{person}{Can Wang}.} \bibinfo{year}{2020}\natexlab{}.
\newblock \showarticletitle{Adaptive graph convolutional recurrent network for traffic forecasting}.
\newblock \bibinfo{journal}{\emph{NeurIPS}} (\bibinfo{year}{2020}).
\newblock


\bibitem[Box and Pierce(1970)]%
        {arima}
\bibfield{author}{\bibinfo{person}{George~EP Box} {and} \bibinfo{person}{David~A Pierce}.} \bibinfo{year}{1970}\natexlab{}.
\newblock \showarticletitle{Distribution of residual autocorrelations in autoregressive-integrated moving average time series models}.
\newblock \bibinfo{journal}{\emph{Journal of the American statistical Association}} (\bibinfo{year}{1970}).
\newblock


\bibitem[Bruna et~al\mbox{.}(2013)]%
        {gcn3}
\bibfield{author}{\bibinfo{person}{Joan Bruna}, \bibinfo{person}{Wojciech Zaremba}, \bibinfo{person}{Arthur Szlam}, {and} \bibinfo{person}{Yann LeCun}.} \bibinfo{year}{2013}\natexlab{}.
\newblock \showarticletitle{Spectral networks and locally connected networks on graphs}.
\newblock \bibinfo{journal}{\emph{arXiv preprint arXiv:1312.6203}} (\bibinfo{year}{2013}).
\newblock


\bibitem[Cao et~al\mbox{.}(2020)]%
        {stemgnn}
\bibfield{author}{\bibinfo{person}{Defu Cao}, \bibinfo{person}{Yujing Wang}, \bibinfo{person}{Juanyong Duan}, \bibinfo{person}{Ce Zhang}, \bibinfo{person}{Xia Zhu}, \bibinfo{person}{Congrui Huang}, \bibinfo{person}{Yunhai Tong}, \bibinfo{person}{Bixiong Xu}, \bibinfo{person}{Jing Bai}, \bibinfo{person}{Jie Tong}, {et~al\mbox{.}}} \bibinfo{year}{2020}\natexlab{}.
\newblock \showarticletitle{Spectral temporal graph neural network for multivariate time-series forecasting}.
\newblock \bibinfo{journal}{\emph{NeurIPS}} (\bibinfo{year}{2020}).
\newblock


\bibitem[Chen et~al\mbox{.}(2019)]%
        {m-rgnn}
\bibfield{author}{\bibinfo{person}{Cen Chen}, \bibinfo{person}{Kenli Li}, \bibinfo{person}{Sin~G Teo}, \bibinfo{person}{Xiaofeng Zou}, \bibinfo{person}{Kang Wang}, \bibinfo{person}{Jie Wang}, {and} \bibinfo{person}{Zeng Zeng}.} \bibinfo{year}{2019}\natexlab{}.
\newblock \showarticletitle{Gated residual recurrent graph neural networks for traffic prediction}. In \bibinfo{booktitle}{\emph{AAAI}}.
\newblock


\bibitem[Chen et~al\mbox{.}(2001)]%
        {pems}
\bibfield{author}{\bibinfo{person}{Chao Chen}, \bibinfo{person}{Karl Petty}, \bibinfo{person}{Alexander Skabardonis}, \bibinfo{person}{Pravin Varaiya}, {and} \bibinfo{person}{Zhanfeng Jia}.} \bibinfo{year}{2001}\natexlab{}.
\newblock \showarticletitle{Freeway performance measurement system: mining loop detector data}.
\newblock \bibinfo{journal}{\emph{Transportation Research Record}} (\bibinfo{year}{2001}).
\newblock


\bibitem[Chen et~al\mbox{.}(2020)]%
        {mra-bgcn}
\bibfield{author}{\bibinfo{person}{Weiqi Chen}, \bibinfo{person}{Ling Chen}, \bibinfo{person}{Yu Xie}, \bibinfo{person}{Wei Cao}, \bibinfo{person}{Yusong Gao}, {and} \bibinfo{person}{Xiaojie Feng}.} \bibinfo{year}{2020}\natexlab{}.
\newblock \showarticletitle{Multi-range attentive bicomponent graph convolutional network for traffic forecasting}. In \bibinfo{booktitle}{\emph{AAAI}}.
\newblock


\bibitem[Chen et~al\mbox{.}(2021)]%
        {zgcnnet}
\bibfield{author}{\bibinfo{person}{Yuzhou Chen}, \bibinfo{person}{Ignacio Segovia}, {and} \bibinfo{person}{Yulia~R Gel}.} \bibinfo{year}{2021}\natexlab{}.
\newblock \showarticletitle{Z-GCNETs: Time Zigzags at Graph Convolutional Networks for Time Series Forecasting}. In \bibinfo{booktitle}{\emph{ICML}}. PMLR.
\newblock


\bibitem[Chung et~al\mbox{.}(2014)]%
        {gru}
\bibfield{author}{\bibinfo{person}{Junyoung Chung}, \bibinfo{person}{Caglar Gulcehre}, \bibinfo{person}{KyungHyun Cho}, {and} \bibinfo{person}{Yoshua Bengio}.} \bibinfo{year}{2014}\natexlab{}.
\newblock \showarticletitle{Empirical evaluation of gated recurrent neural networks on sequence modeling}.
\newblock \bibinfo{journal}{\emph{arXiv preprint arXiv:1412.3555}} (\bibinfo{year}{2014}).
\newblock


\bibitem[Dauphin et~al\mbox{.}(2017)]%
        {gate}
\bibfield{author}{\bibinfo{person}{Yann~N Dauphin}, \bibinfo{person}{Angela Fan}, \bibinfo{person}{Michael Auli}, {and} \bibinfo{person}{David Grangier}.} \bibinfo{year}{2017}\natexlab{}.
\newblock \showarticletitle{Language modeling with gated convolutional networks}. In \bibinfo{booktitle}{\emph{ICML}}. PMLR.
\newblock


\bibitem[Defferrard et~al\mbox{.}(2016)]%
        {gcn2}
\bibfield{author}{\bibinfo{person}{Micha{\"e}l Defferrard}, \bibinfo{person}{Xavier Bresson}, {and} \bibinfo{person}{Pierre Vandergheynst}.} \bibinfo{year}{2016}\natexlab{}.
\newblock \showarticletitle{Convolutional neural networks on graphs with fast localized spectral filtering}.
\newblock \bibinfo{journal}{\emph{NeurIPS}}  \bibinfo{volume}{29} (\bibinfo{year}{2016}).
\newblock


\bibitem[Fang et~al\mbox{.}(2021)]%
        {stgode}
\bibfield{author}{\bibinfo{person}{Zheng Fang}, \bibinfo{person}{Qingqing Long}, \bibinfo{person}{Guojie Song}, {and} \bibinfo{person}{Kunqing Xie}.} \bibinfo{year}{2021}\natexlab{}.
\newblock \showarticletitle{Spatial-temporal graph ode networks for traffic flow forecasting}. In \bibinfo{booktitle}{\emph{KDD}}.
\newblock


\bibitem[Guo et~al\mbox{.}(2021)]%
        {hgcn}
\bibfield{author}{\bibinfo{person}{Kan Guo}, \bibinfo{person}{Yongli Hu}, \bibinfo{person}{Yanfeng Sun}, \bibinfo{person}{Sean Qian}, \bibinfo{person}{Junbin Gao}, {and} \bibinfo{person}{Baocai Yin}.} \bibinfo{year}{2021}\natexlab{}.
\newblock \showarticletitle{Hierarchical graph convolution networks for traffic forecasting}. In \bibinfo{booktitle}{\emph{AAAI}}.
\newblock


\bibitem[Guo et~al\mbox{.}(2019)]%
        {astgcn}
\bibfield{author}{\bibinfo{person}{Shengnan Guo}, \bibinfo{person}{Youfang Lin}, \bibinfo{person}{Ning Feng}, \bibinfo{person}{Chao Song}, {and} \bibinfo{person}{Huaiyu Wan}.} \bibinfo{year}{2019}\natexlab{}.
\newblock \showarticletitle{Attention based spatial-temporal graph convolutional networks for traffic flow forecasting}. In \bibinfo{booktitle}{\emph{AAAI}}.
\newblock


\bibitem[Hamilton(2020)]%
        {hamilton2020time}
\bibfield{author}{\bibinfo{person}{James~Douglas Hamilton}.} \bibinfo{year}{2020}\natexlab{}.
\newblock \bibinfo{booktitle}{\emph{Time series analysis}}.
\newblock \bibinfo{publisher}{Princeton university press}.
\newblock


\bibitem[Han et~al\mbox{.}(2021)]%
        {dynamic_time_3}
\bibfield{author}{\bibinfo{person}{Liangzhe Han}, \bibinfo{person}{Bowen Du}, \bibinfo{person}{Leilei Sun}, \bibinfo{person}{Yanjie Fu}, \bibinfo{person}{Yisheng Lv}, {and} \bibinfo{person}{Hui Xiong}.} \bibinfo{year}{2021}\natexlab{}.
\newblock \showarticletitle{Dynamic and multi-faceted spatio-temporal deep learning for traffic speed forecasting}. In \bibinfo{booktitle}{\emph{Proceedings of the 27th ACM SIGKDD conference on knowledge discovery \& data mining}}. \bibinfo{pages}{547--555}.
\newblock


\bibitem[Hochreiter and Schmidhuber(1997)]%
        {lstm}
\bibfield{author}{\bibinfo{person}{Sepp Hochreiter} {and} \bibinfo{person}{J{\"u}rgen Schmidhuber}.} \bibinfo{year}{1997}\natexlab{}.
\newblock \showarticletitle{Long short-term memory}.
\newblock \bibinfo{journal}{\emph{Neural computation}} \bibinfo{volume}{9}, \bibinfo{number}{8} (\bibinfo{year}{1997}), \bibinfo{pages}{1735--1780}.
\newblock


\bibitem[Huang et~al\mbox{.}(2020)]%
        {lsgcn}
\bibfield{author}{\bibinfo{person}{Rongzhou Huang}, \bibinfo{person}{Chuyin Huang}, \bibinfo{person}{Yubao Liu}, \bibinfo{person}{Genan Dai}, {and} \bibinfo{person}{Weiyang Kong}.} \bibinfo{year}{2020}\natexlab{}.
\newblock \showarticletitle{LSGCN: Long Short-Term Traffic Prediction with Graph Convolutional Networks.}. In \bibinfo{booktitle}{\emph{IJCAI}}. \bibinfo{pages}{2355--2361}.
\newblock


\bibitem[Jang et~al\mbox{.}(2016)]%
        {gumbel2}
\bibfield{author}{\bibinfo{person}{Eric Jang}, \bibinfo{person}{Shixiang Gu}, {and} \bibinfo{person}{Ben Poole}.} \bibinfo{year}{2016}\natexlab{}.
\newblock \showarticletitle{Categorical reparameterization with gumbel-softmax}.
\newblock \bibinfo{journal}{\emph{arXiv preprint arXiv:1611.01144}} (\bibinfo{year}{2016}).
\newblock


\bibitem[Kipf and Welling(2016)]%
        {gcn}
\bibfield{author}{\bibinfo{person}{Thomas~N Kipf} {and} \bibinfo{person}{Max Welling}.} \bibinfo{year}{2016}\natexlab{}.
\newblock \showarticletitle{Semi-supervised classification with graph convolutional networks}.
\newblock \bibinfo{journal}{\emph{arXiv preprint arXiv:1609.02907}} (\bibinfo{year}{2016}).
\newblock


\bibitem[Lan et~al\mbox{.}(2022)]%
        {dstagnn}
\bibfield{author}{\bibinfo{person}{Shiyong Lan}, \bibinfo{person}{Yitong Ma}, \bibinfo{person}{Weikang Huang}, \bibinfo{person}{Wenwu Wang}, \bibinfo{person}{Hongyu Yang}, {and} \bibinfo{person}{Pyang Li}.} \bibinfo{year}{2022}\natexlab{}.
\newblock \showarticletitle{DSTAGNN: Dynamic Spatial-Temporal Aware Graph Neural Network for Traffic Flow Forecasting}. In \bibinfo{booktitle}{\emph{International Conference on Machine Learning}}. PMLR, \bibinfo{pages}{11906--11917}.
\newblock


\bibitem[Li et~al\mbox{.}(2023)]%
        {dynamic_time_1}
\bibfield{author}{\bibinfo{person}{Fuxian Li}, \bibinfo{person}{Jie Feng}, \bibinfo{person}{Huan Yan}, \bibinfo{person}{Guangyin Jin}, \bibinfo{person}{Fan Yang}, \bibinfo{person}{Funing Sun}, \bibinfo{person}{Depeng Jin}, {and} \bibinfo{person}{Yong Li}.} \bibinfo{year}{2023}\natexlab{}.
\newblock \showarticletitle{Dynamic graph convolutional recurrent network for traffic prediction: Benchmark and solution}.
\newblock \bibinfo{journal}{\emph{ACM Transactions on Knowledge Discovery from Data}} \bibinfo{volume}{17}, \bibinfo{number}{1} (\bibinfo{year}{2023}), \bibinfo{pages}{1--21}.
\newblock


\bibitem[Li and Zhu(2021)]%
        {stfgnn}
\bibfield{author}{\bibinfo{person}{Mengzhang Li} {and} \bibinfo{person}{Zhanxing Zhu}.} \bibinfo{year}{2021}\natexlab{}.
\newblock \showarticletitle{Spatial-temporal fusion graph neural networks for traffic flow forecasting}. In \bibinfo{booktitle}{\emph{AAAI}}.
\newblock


\bibitem[Li et~al\mbox{.}(2017)]%
        {dcrnn}
\bibfield{author}{\bibinfo{person}{Yaguang Li}, \bibinfo{person}{Rose Yu}, \bibinfo{person}{Cyrus Shahabi}, {and} \bibinfo{person}{Yan Liu}.} \bibinfo{year}{2017}\natexlab{}.
\newblock \showarticletitle{Diffusion convolutional recurrent neural network: Data-driven traffic forecasting}.
\newblock \bibinfo{journal}{\emph{arXiv preprint arXiv:1707.01926}} (\bibinfo{year}{2017}).
\newblock


\bibitem[Lu et~al\mbox{.}(2020)]%
        {stag-gcn}
\bibfield{author}{\bibinfo{person}{Bin Lu}, \bibinfo{person}{Xiaoying Gan}, \bibinfo{person}{Haiming Jin}, \bibinfo{person}{Luoyi Fu}, {and} \bibinfo{person}{Haisong Zhang}.} \bibinfo{year}{2020}\natexlab{}.
\newblock \showarticletitle{Spatiotemporal adaptive gated graph convolution network for urban traffic flow forecasting}. In \bibinfo{booktitle}{\emph{CIKM}}.
\newblock


\bibitem[Maddison et~al\mbox{.}(2016)]%
        {gumbel}
\bibfield{author}{\bibinfo{person}{Chris~J Maddison}, \bibinfo{person}{Andriy Mnih}, {and} \bibinfo{person}{Yee~Whye Teh}.} \bibinfo{year}{2016}\natexlab{}.
\newblock \showarticletitle{The concrete distribution: A continuous relaxation of discrete random variables}.
\newblock \bibinfo{journal}{\emph{arXiv preprint arXiv:1611.00712}} (\bibinfo{year}{2016}).
\newblock


\bibitem[Park et~al\mbox{.}(2020)]%
        {st-gart}
\bibfield{author}{\bibinfo{person}{Cheonbok Park}, \bibinfo{person}{Chunggi Lee}, \bibinfo{person}{Hyojin Bahng}, \bibinfo{person}{Yunwon Tae}, \bibinfo{person}{Seungmin Jin}, \bibinfo{person}{Kihwan Kim}, \bibinfo{person}{Sungahn Ko}, {and} \bibinfo{person}{Jaegul Choo}.} \bibinfo{year}{2020}\natexlab{}.
\newblock \showarticletitle{ST-GRAT: A novel spatio-temporal graph attention networks for accurately forecasting dynamically changing road speed}. In \bibinfo{booktitle}{\emph{CIKM}}.
\newblock


\bibitem[Shang et~al\mbox{.}(2021)]%
        {gts}
\bibfield{author}{\bibinfo{person}{Chao Shang}, \bibinfo{person}{Jie Chen}, {and} \bibinfo{person}{Jinbo Bi}.} \bibinfo{year}{2021}\natexlab{}.
\newblock \showarticletitle{Discrete graph structure learning for forecasting multiple time series}.
\newblock \bibinfo{journal}{\emph{arXiv preprint arXiv:2101.06861}} (\bibinfo{year}{2021}).
\newblock


\bibitem[Shao et~al\mbox{.}(2022)]%
        {dynamic_time_2}
\bibfield{author}{\bibinfo{person}{Zezhi Shao}, \bibinfo{person}{Zhao Zhang}, \bibinfo{person}{Wei Wei}, \bibinfo{person}{Fei Wang}, \bibinfo{person}{Yongjun Xu}, \bibinfo{person}{Xin Cao}, {and} \bibinfo{person}{Christian~S Jensen}.} \bibinfo{year}{2022}\natexlab{}.
\newblock \showarticletitle{Decoupled dynamic spatial-temporal graph neural network for traffic forecasting}.
\newblock \bibinfo{journal}{\emph{arXiv preprint arXiv:2206.09112}} (\bibinfo{year}{2022}).
\newblock


\bibitem[Song et~al\mbox{.}(2020)]%
        {stsgcn}
\bibfield{author}{\bibinfo{person}{Chao Song}, \bibinfo{person}{Youfang Lin}, \bibinfo{person}{Shengnan Guo}, {and} \bibinfo{person}{Huaiyu Wan}.} \bibinfo{year}{2020}\natexlab{}.
\newblock \showarticletitle{Spatial-temporal synchronous graph convolutional networks: A new framework for spatial-temporal network data forecasting}. In \bibinfo{booktitle}{\emph{AAAI}}.
\newblock


\bibitem[Sutskever et~al\mbox{.}(2014)]%
        {fclstm}
\bibfield{author}{\bibinfo{person}{Ilya Sutskever}, \bibinfo{person}{Oriol Vinyals}, {and} \bibinfo{person}{Quoc~V Le}.} \bibinfo{year}{2014}\natexlab{}.
\newblock \showarticletitle{Sequence to sequence learning with neural networks}.
\newblock \bibinfo{journal}{\emph{NeurIPS}} (\bibinfo{year}{2014}).
\newblock


\bibitem[Tedjopurnomo et~al\mbox{.}(2020)]%
        {tedjopurnomo2020survey}
\bibfield{author}{\bibinfo{person}{David~Alexander Tedjopurnomo}, \bibinfo{person}{Zhifeng Bao}, \bibinfo{person}{Baihua Zheng}, \bibinfo{person}{Farhana Choudhury}, {and} \bibinfo{person}{AK Qin}.} \bibinfo{year}{2020}\natexlab{}.
\newblock \showarticletitle{A survey on modern deep neural network for traffic prediction: Trends, methods and challenges}.
\newblock \bibinfo{journal}{\emph{TKDE}} (\bibinfo{year}{2020}).
\newblock


\bibitem[Thakoor et~al\mbox{.}(2022)]%
        {thakoor2021large}
\bibfield{author}{\bibinfo{person}{Shantanu Thakoor}, \bibinfo{person}{Corentin Tallec}, \bibinfo{person}{Mohammad~Gheshlaghi Azar}, \bibinfo{person}{Mehdi Azabou}, \bibinfo{person}{Eva~L Dyer}, \bibinfo{person}{Remi Munos}, \bibinfo{person}{Petar Veli{\v{c}}kovi{\'c}}, {and} \bibinfo{person}{Michal Valko}.} \bibinfo{year}{2022}\natexlab{}.
\newblock \showarticletitle{Large-scale representation learning on graphs via bootstrapping}.
\newblock \bibinfo{journal}{\emph{ICLR}} (\bibinfo{year}{2022}).
\newblock


\bibitem[Veli{\v{c}}kovi{\'c} et~al\mbox{.}(2017)]%
        {gat}
\bibfield{author}{\bibinfo{person}{Petar Veli{\v{c}}kovi{\'c}}, \bibinfo{person}{Guillem Cucurull}, \bibinfo{person}{Arantxa Casanova}, \bibinfo{person}{Adriana Romero}, \bibinfo{person}{Pietro Lio}, {and} \bibinfo{person}{Yoshua Bengio}.} \bibinfo{year}{2017}\natexlab{}.
\newblock \showarticletitle{Graph attention networks}.
\newblock \bibinfo{journal}{\emph{arXiv preprint arXiv:1710.10903}} (\bibinfo{year}{2017}).
\newblock


\bibitem[Wu et~al\mbox{.}(2004)]%
        {svr}
\bibfield{author}{\bibinfo{person}{Chun-Hsin Wu}, \bibinfo{person}{Jan-Ming Ho}, {and} \bibinfo{person}{Der-Tsai Lee}.} \bibinfo{year}{2004}\natexlab{}.
\newblock \showarticletitle{Travel-time prediction with support vector regression}.
\newblock \bibinfo{journal}{\emph{IEEE transactions on intelligent transportation systems}} (\bibinfo{year}{2004}).
\newblock


\bibitem[Wu et~al\mbox{.}(2019)]%
        {wavenet}
\bibfield{author}{\bibinfo{person}{Zonghan Wu}, \bibinfo{person}{Shirui Pan}, \bibinfo{person}{Guodong Long}, \bibinfo{person}{Jing Jiang}, {and} \bibinfo{person}{Chengqi Zhang}.} \bibinfo{year}{2019}\natexlab{}.
\newblock \showarticletitle{Graph wavenet for deep spatial-temporal graph modeling}.
\newblock \bibinfo{journal}{\emph{arXiv preprint arXiv:1906.00121}} (\bibinfo{year}{2019}).
\newblock


\bibitem[Xie et~al\mbox{.}(2020)]%
        {digc-net}
\bibfield{author}{\bibinfo{person}{Qinge Xie}, \bibinfo{person}{Tiancheng Guo}, \bibinfo{person}{Yang Chen}, \bibinfo{person}{Yu Xiao}, \bibinfo{person}{Xin Wang}, {and} \bibinfo{person}{Ben~Y Zhao}.} \bibinfo{year}{2020}\natexlab{}.
\newblock \showarticletitle{Deep graph convolutional networks for incident-driven traffic speed prediction}. In \bibinfo{booktitle}{\emph{CIKM}}.
\newblock


\bibitem[Yao et~al\mbox{.}(2018a)]%
        {rnncnn2}
\bibfield{author}{\bibinfo{person}{Huaxiu Yao}, \bibinfo{person}{Xianfeng Tang}, \bibinfo{person}{Hua Wei}, \bibinfo{person}{Guanjie Zheng}, \bibinfo{person}{Yanwei Yu}, {and} \bibinfo{person}{Zhenhui Li}.} \bibinfo{year}{2018}\natexlab{a}.
\newblock \showarticletitle{Modeling spatial-temporal dynamics for traffic prediction}.
\newblock \bibinfo{journal}{\emph{arXiv preprint arXiv:1803.01254}} (\bibinfo{year}{2018}).
\newblock


\bibitem[Yao et~al\mbox{.}(2018b)]%
        {rnncnn3}
\bibfield{author}{\bibinfo{person}{Huaxiu Yao}, \bibinfo{person}{Fei Wu}, \bibinfo{person}{Jintao Ke}, \bibinfo{person}{Xianfeng Tang}, \bibinfo{person}{Yitian Jia}, \bibinfo{person}{Siyu Lu}, \bibinfo{person}{Pinghua Gong}, \bibinfo{person}{Jieping Ye}, {and} \bibinfo{person}{Zhenhui Li}.} \bibinfo{year}{2018}\natexlab{b}.
\newblock \showarticletitle{Deep multi-view spatial-temporal network for taxi demand prediction}. In \bibinfo{booktitle}{\emph{AAAI}}.
\newblock


\bibitem[Ye et~al\mbox{.}(2022)]%
        {dynamic_time_4}
\bibfield{author}{\bibinfo{person}{Junchen Ye}, \bibinfo{person}{Zihan Liu}, \bibinfo{person}{Bowen Du}, \bibinfo{person}{Leilei Sun}, \bibinfo{person}{Weimiao Li}, \bibinfo{person}{Yanjie Fu}, {and} \bibinfo{person}{Hui Xiong}.} \bibinfo{year}{2022}\natexlab{}.
\newblock \showarticletitle{Learning the evolutionary and multi-scale graph structure for multivariate time series forecasting}. In \bibinfo{booktitle}{\emph{Proceedings of the 28th ACM SIGKDD conference on knowledge discovery and data mining}}. \bibinfo{pages}{2296--2306}.
\newblock


\bibitem[Yu et~al\mbox{.}(2017)]%
        {stgcn}
\bibfield{author}{\bibinfo{person}{Bing Yu}, \bibinfo{person}{Haoteng Yin}, {and} \bibinfo{person}{Zhanxing Zhu}.} \bibinfo{year}{2017}\natexlab{}.
\newblock \showarticletitle{Spatio-temporal graph convolutional networks: A deep learning framework for traffic forecasting}.
\newblock \bibinfo{journal}{\emph{arXiv preprint arXiv:1709.04875}} (\bibinfo{year}{2017}).
\newblock


\bibitem[Zhang et~al\mbox{.}(2018a)]%
        {gaan}
\bibfield{author}{\bibinfo{person}{Jiani Zhang}, \bibinfo{person}{Xingjian Shi}, \bibinfo{person}{Junyuan Xie}, \bibinfo{person}{Hao Ma}, \bibinfo{person}{Irwin King}, {and} \bibinfo{person}{Dit-Yan Yeung}.} \bibinfo{year}{2018}\natexlab{a}.
\newblock \showarticletitle{Gaan: Gated attention networks for learning on large and spatiotemporal graphs}.
\newblock \bibinfo{journal}{\emph{arXiv preprint arXiv:1803.07294}} (\bibinfo{year}{2018}).
\newblock


\bibitem[Zhang et~al\mbox{.}(2018b)]%
        {rnncnn}
\bibfield{author}{\bibinfo{person}{Junbo Zhang}, \bibinfo{person}{Yu Zheng}, \bibinfo{person}{Dekang Qi}, \bibinfo{person}{Ruiyuan Li}, \bibinfo{person}{Xiuwen Yi}, {and} \bibinfo{person}{Tianrui Li}.} \bibinfo{year}{2018}\natexlab{b}.
\newblock \showarticletitle{Predicting citywide crowd flows using deep spatio-temporal residual networks}.
\newblock \bibinfo{journal}{\emph{Artificial Intelligence}} (\bibinfo{year}{2018}).
\newblock


\bibitem[Zheng et~al\mbox{.}(2020)]%
        {gman}
\bibfield{author}{\bibinfo{person}{Chuanpan Zheng}, \bibinfo{person}{Xiaoliang Fan}, \bibinfo{person}{Cheng Wang}, {and} \bibinfo{person}{Jianzhong Qi}.} \bibinfo{year}{2020}\natexlab{}.
\newblock \showarticletitle{Gman: A graph multi-attention network for traffic prediction}. In \bibinfo{booktitle}{\emph{AAAI}}.
\newblock


\bibitem[Zhu et~al\mbox{.}(2021b)]%
        {gsl}
\bibfield{author}{\bibinfo{person}{Yanqiao Zhu}, \bibinfo{person}{Weizhi Xu}, \bibinfo{person}{Jinghao Zhang}, \bibinfo{person}{Yuanqi Du}, \bibinfo{person}{Jieyu Zhang}, \bibinfo{person}{Qiang Liu}, \bibinfo{person}{Carl Yang}, {and} \bibinfo{person}{Shu Wu}.} \bibinfo{year}{2021}\natexlab{b}.
\newblock \showarticletitle{A survey on graph structure learning: Progress and opportunities}.
\newblock \bibinfo{journal}{\emph{arXiv e-prints}} (\bibinfo{year}{2021}), \bibinfo{pages}{arXiv--2103}.
\newblock


\bibitem[Zhu et~al\mbox{.}(2020)]%
        {zhu2020deep}
\bibfield{author}{\bibinfo{person}{Yanqiao Zhu}, \bibinfo{person}{Yichen Xu}, \bibinfo{person}{Feng Yu}, \bibinfo{person}{Qiang Liu}, \bibinfo{person}{Shu Wu}, {and} \bibinfo{person}{Liang Wang}.} \bibinfo{year}{2020}\natexlab{}.
\newblock \showarticletitle{Deep graph contrastive representation learning}.
\newblock \bibinfo{journal}{\emph{arXiv preprint arXiv:2006.04131}} (\bibinfo{year}{2020}).
\newblock


\bibitem[Zhu et~al\mbox{.}(2021a)]%
        {zhu2021graph}
\bibfield{author}{\bibinfo{person}{Yanqiao Zhu}, \bibinfo{person}{Yichen Xu}, \bibinfo{person}{Feng Yu}, \bibinfo{person}{Qiang Liu}, \bibinfo{person}{Shu Wu}, {and} \bibinfo{person}{Liang Wang}.} \bibinfo{year}{2021}\natexlab{a}.
\newblock \showarticletitle{Graph contrastive learning with adaptive augmentation}. In \bibinfo{booktitle}{\emph{Proceedings of the Web Conference 2021}}. \bibinfo{pages}{2069--2080}.
\newblock


\end{thebibliography}

\end{document}